\documentclass[10pt,twocolumn,letterpaper]{article}

\usepackage[pagenumbers]{cvpr} 

\usepackage{marvosym} 
\usepackage[hypcap=false]{caption}
\usepackage{xcolor}
\usepackage{colortbl}
\usepackage{multirow}
\usepackage{times}
\usepackage{epsfig}
\usepackage{amsmath}
\usepackage{amssymb}
\usepackage{graphicx}
\usepackage{booktabs}
\usepackage{tabularx}
\usepackage{makecell}
\usepackage{caption}
\usepackage{float}
\usepackage{tcolorbox}
\usepackage[accsupp]{axessibility}

\tcbuselibrary{skins}

\newcommand{\best}[1]{\textbf{#1}}
\newcommand{\second}[1]{\underline{#1}}

\definecolor{cvprblue}{rgb}{0.21,0.49,0.74}
\usepackage[pagebackref,breaklinks,colorlinks,allcolors=cvprblue]{hyperref}

\def\paperID{37192} 
\def\confName{CVPR}
\def\confYear{2026}

\newcommand{\name}{ConsistCompose\xspace}
\title{\name: Unified Multimodal Layout Control for Image Composition}

\author{
  Xuanke Shi \quad
  Boxuan Li \quad
  Xiaoyang Han \\
  Zhongang Cai \quad
  Lei Yang\footnotemark[1] \quad
  Quan Wang\footnotemark[1] \quad
  Dahua Lin\footnotemark[1] \\
  SenseTime Research
}

\begin{document}

\twocolumn[{
    \centering
    \maketitle
    \vspace{-20pt}
    \includegraphics[width=0.94\linewidth]{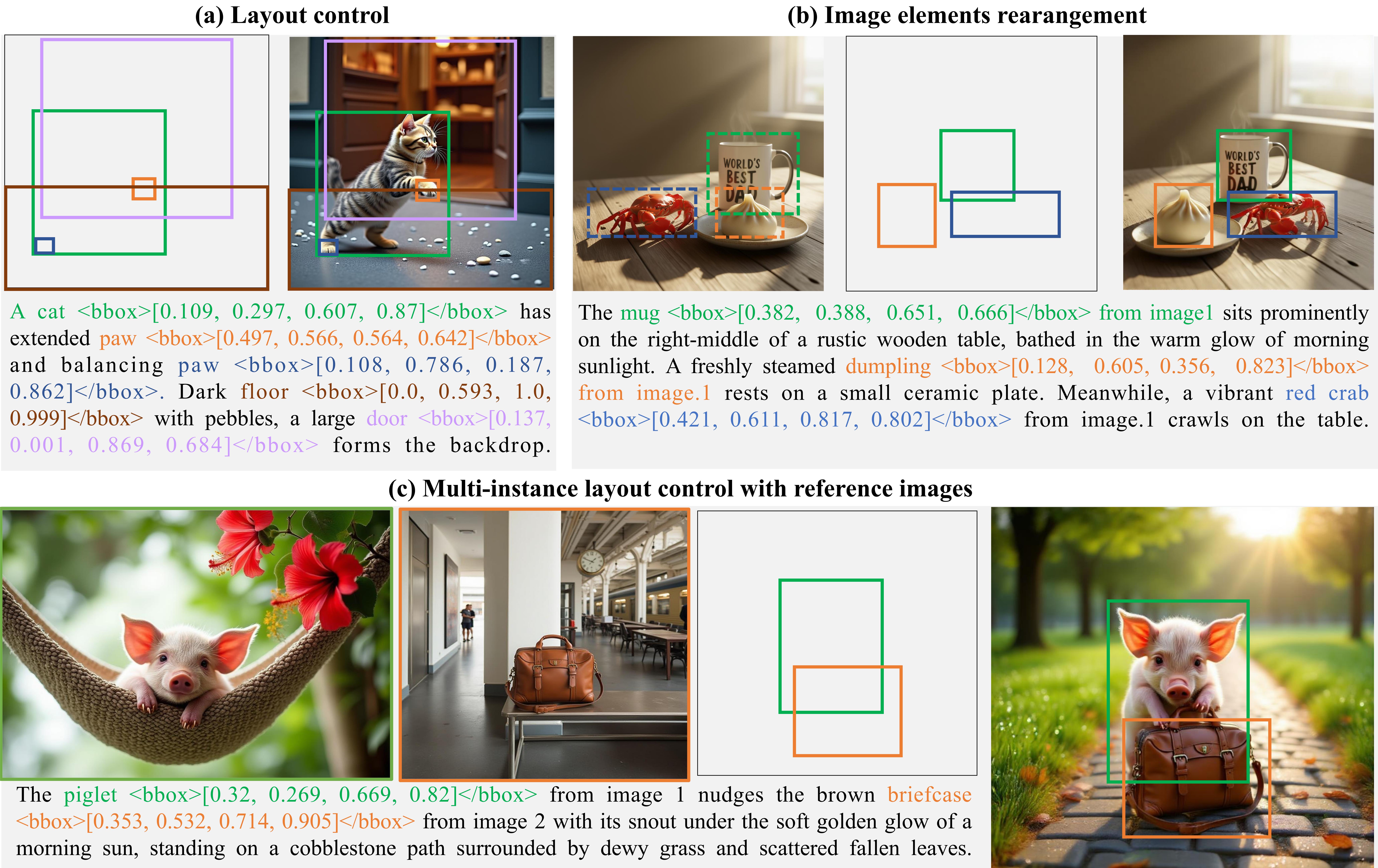}
    \vspace{-4pt}
    \captionof{figure}{Key Results of ConsistCompose: Advancing Layout-Controllable Multi-Instance Generation. (a) Layout-Grounded Text-to-Image Generation with Faithful Prompt Alignment and High Precision. (b) Content-Preserving, Flexible Rearrangement of Image Elements. (c) Multi-Instance Layout Control for Structured Complex Scene Generation with Identity Preservation.}
    \label{fig:banner}
    \vspace{2pt}
}]

\maketitle

\begingroup
\renewcommand{\thefootnote}{\fnsymbol{footnote}}
\phantomsection
\footnotetext[1]{Corresponding author.}
\endgroup

\begin{abstract}
Unified multimodal models that couple visual understanding with image generation have advanced rapidly, yet most systems still focus on visual grounding—aligning language with image regions—while their generative counterpart, \emph{linguistic-embedded layout-grounded generation} (LELG) for layout-controllable multi-instance generation, remains underexplored and limits precise compositional control. We present ConsistCompose, a unified multimodal framework that embeds layout coordinates directly into language prompts, enabling layout-controlled multi-instance image generation from Interleaved Image-Text within a single generative interface. We further construct ConsistCompose3M, a 3.4M multi-instance generation dataset  with layout and identity annotations (2.6M text-guided and 0.8M image-guided data pairs) that provides large-scale supervision for layout-conditioned generation. Within this framework, LELG is instantiated through instance–coordinate binding prompts and coordinate-aware classifier-free guidance, which translate linguistic layout cues into precise spatial control without task-specific branches. Experiments on COCO-Position and MS-Bench show that ConsistCompose substantially improves spatial accuracy over layout-controlled baselines while preserving identity fidelity and competitive general multimodal understanding, establishing a unified paradigm for layout-controllable multimodal image generation.
\end{abstract}    
\section{Introduction}
\label{sec:intro}

Recent advances in diffusion models~\cite{ho2020denoising} and large-scale vision–language training have substantially improved the quality and versatility of image generation. Systems such as DALL-E~\cite{ramesh2021zero} and Stable Diffusion~\cite{rombach2022high} demonstrate impressive photorealism, while recent unified multimodal models~\cite{xie2024show, zhou2024transfusion, wu2025janus, zhang2025unified, pan2025transfer, chen2025janus, deng2025emerging} integrate perception, reasoning, and generation into a single architecture. Despite their broad capabilities, these unified models remain predominantly focused on visual understanding—especially grounding~\cite{fu2025llmdet, wang2025learning, jiang2025detect}—and provide limited support for generating images that satisfy explicit spatial constraints, which is a prerequisite for controllable and compositional scene synthesis.

Layout-controlled generation has emerged as an important direction for enabling spatially grounded image synthesis. However, existing approaches present fundamental obstacles to unification. Diffusion-based methods typically rely on layout–image fusion modules or region-aware U-Net modifications~\cite{li2023gligen, wang2024instancediffusion, Zhou2024MIGCMG, zhou2024migc++, wang2025msdiffusion, peng2025muse}, making their architectures tightly coupled to layout-specific designs and difficult to reconcile with Transformer-based generative frameworks. Multimodal or autoregressive models such as LayoutSAM~\cite{zhang2025creatilayout}, HiCo~\cite{bocheng2024hico}, SMARLI~\cite{zheng2025layout}, and PlanGen~\cite{he2025plangen} attempt to incorporate layout either as an independent modality or through structured spatial tokens, and some explore unified backbones for layout planning and layout-to-image generation. Nevertheless, these systems remain restricted to layout-centric tasks and have not demonstrated compatibility with broader multimodal abilities such as visual reasoning~\cite{shao2024visual}, image understanding, reference grounding, or editing. Moreover, most prior work focuses solely on text-conditioned layout control and seldom considers the more challenging setting of multi-reference, identity-preserving multi-instance composition. This limitation is further compounded by the lack of large-scale datasets that jointly provide instance-level layouts, subject correspondences, and diverse multi-instance configurations, limiting progress toward truly unified layout-aware multimodal systems.

To overcome these limitations, we take a different perspective: instead of designing specialized spatial planners or layout encoders, we embed layout constraints directly into the language interface. When layout semantics are represented as part of the text, a multimodal Transformer can learn spatial grounding through the same interleaved modeling used for perception and reasoning, without any architecture-specific modifications. This leads to a simple but principled paradigm, which we refer to as Linguistic-Embedded Layout-Grounded Generation (LELG). Unlike prior layout-specific pipelines, LELG encodes layout as explicit textual control tokens in the linguistic stream, enabling seamless integration of spatial constraints into unified multimodal modeling.

Building on this paradigm, we propose ConsistCompose, a unified multimodal generation framework that jointly supports layout-grounded text-to-image synthesis, multi-reference identity-consistent multi-instance composition, and general multimodal understanding within a single model. Spatial constraints are represented as succinct coordinate expressions in text, allowing the model to bind each subject identity and its designated position through the shared token space governing both understanding and generation. A coordinate-aware classifier-free guidance mechanism further enhances spatial fidelity during sampling, without altering the backbone architecture. This formulation naturally combines spatial reasoning, identity preservation, and multimodal generation—capabilities that previous systems address only in isolation.

To train this framework at scale, we construct ConsistCompose3M, a 3.4M-sample multimodal dataset that includes both layout-grounded text-to-image examples and large-scale multi-reference identity-conditioned scenes. The dataset provides instance-level layout annotations, explicit subject correspondences, and diverse compositional patterns tailored for unified modeling, rather than for layout-specific pipelines.

Our experiments show that ConsistCompose delivers state-of-the-art layout controllability, achieving a 7.2\% gain in layout IoU and a 13.7\% AP improvement on COCO-Position. Beyond COCO-Position, it exhibits strong identity preservation under multi-reference layouts on MS-Bench~\cite{wang2025msdiffusion}, while maintaining MMMU and MMBench performance on par with its Bagel backbone. These results confirm that explicit spatial control can be incorporated into unified multimodal generation without sacrificing general capabilities.

Our contributions are threefold:  
\begin{itemize}
\item We introduce LELG, a language-driven paradigm that integrates explicit spatial constraints into unified multimodal understanding–generation modeling.  
\item We propose ConsistCompose, a unified multimodal framework that consolidates layout-grounded text-to-image generation, multi-reference identity-preserving composition, and general multimodal understanding and generation.
\item We construct ConsistCompose3M, a 3.4M-sample dataset that supplies the spatial and identity supervision required for unified layout-aware multimodal training.
\end{itemize}
\section{Related Work}
\label{sec:related}

\noindent\textbf{Text-to-Image Generation.}
Text-to-image (T2I) generation has progressed rapidly with the development of diffusion models~\cite{ho2020denoising, podell2023sdxl, ramesh2021zero, rombach2022high, esser2024scaling, cao2025hunyuanimage, Wu2025QwenImageTR}, enabling high-fidelity and semantically coherent image generation across diverse domains.
However, purely text-driven guidance offers limited control over spatial composition and structural layout, restricting its applicability in scenarios that require precise object placement, poses, or scene configurations.

\noindent\textbf{Layout-Controlled Image Generation.}
Beyond dense-structure conditioning methods such as ControlNet~\cite{zhang2023adding}, T2I-Adapter~\cite{mou2024t2i}, and related approaches~\cite{voynov2023sketch, tan2025ominicontrol, xiao2025omnigen}, which enforce geometric priors via edges or segmentation maps, layout-based approaches~\cite{li2023gligen, wang2024instancediffusion, bocheng2024hico, zhang2025creatilayout, Zhou2024MIGCMG, zhou2024migc++} offer a more flexible means of spatial control.
GLIGEN~\cite{li2023gligen} introduces bounding-box constraints through gated Transformer layers for open-set generalization, while InstanceDiffusion~\cite{wang2024instancediffusion} achieves instance-level control via multimodal input fusion.
MIGC~\cite{Zhou2024MIGCMG, zhou2024migc++} mitigates attribute leakage and spatial misalignment with a divide-and-conquer attention mechanism, and CreatiLayout~\cite{zhang2025creatilayout} treats layout as an independent modality, employing SiamLayout to balance text–layout interactions for improved consistency.
Recent studies further extend this line toward \textit{multi-instance generation} guided by both layout and visual references:
MUSE~\cite{peng2025muse} fuses layout and instance information via concatenated Cross-Attention; MS-Diffusion~\cite{wang2025msdiffusion} introduces a Grounding Resampler with Multi-Subject Cross-Attention for coherent multi-subject synthesis; and ContextGen~\cite{xu2025contextgen} enhances spatial–appearance alignment through Contextual Layout Anchoring and Identity Consistency Attention.
Despite their progress, these approaches remain fragmented and highly task-specific, often relying on auxiliary fusion modules or handcrafted attention mechanisms.
Such designs limit scalability and hinder integration into unified multimodal frameworks.
In contrast, our work explores LELG under a unified multimodal paradigm, embedding layout semantics directly into language prompts to achieve consistent, layout-aware, and identity-preserving multi-instance generation.

\noindent\textbf{Unified Multimodal Generation.}
The recent rise of multimodal large language models (MLLMs) has enabled unified architectures that perform both \textit{understanding} and \textit{generation} within a single framework~\cite{zhang2025unified}.
In parallel, diffusion-based models have evolved toward \textit{unified multimodal generation}, where a single model supports diverse generation tasks under shared objectives~\cite{tan2025ominicontrol, tan2025ominicontrol2, xiao2025omnigen, wu2025omnigen2explorationadvancedmultimodal}.
The release of GPT-4o~\cite{hurst2024gpt} further underscores the potential of such cross-modal unification.
Early efforts, including Chameleon~\cite{team2024chameleon} and Emu2~\cite{Emu2}, extend language modeling to multimodal contexts by treating images as token sequences and applying next-token prediction across modalities.
Emu3~\cite{wang2024emu3} enhances this design via discrete VQ-VAE tokenization, enabling unified autoregressive modeling within a shared token space.
Hybrid frameworks such as Show-O~\cite{xie2024show} and Transfusion~\cite{zhou2024transfusion} combine Transformers and diffusion within a joint pipeline to integrate token prediction and denoising.
Janus~\cite{wu2025janus} and Janus Pro~\cite{chen2025janus} reveal that a single visual encoder struggles to balance perception and generation, while MetaQuery~\cite{pan2025transfer} and BLIP-3o~\cite{chen2025blip3} encode conditional information via fixed query tokens, which limits representational capacity and leads to information loss for complex prompts.
OmniGen2~\cite{wu2025omnigen2explorationadvancedmultimodal} alleviates this issue by directly leveraging interleaved multimodal hidden states from MLLMs as diffusion inputs, avoiding token-level compression.

Inspired by the MMDiT mechanism, recent systems such as Bagel~\cite{deng2025emerging} and Mogao~\cite{liao2025mogao} adopt a \textit{deep-fusion} architecture that jointly processes textual and visual tokens within shared self-attention layers, achieving fine-grained modality interaction and improved semantic alignment.
Building upon this evolution, we further investigate the MoT architecture and extend its deep-fusion mechanism for controllable multi-instance image generation.
Our framework targets the key challenge of layout-controlled multimodal generation—ensuring spatial precision, compositional consistency, and identity preservation—through LELG, which embeds layout semantics and instance identities directly within the unified vision-language interface under a single generative paradigm.
\section{Method}
\label{sec:method}

\begin{figure*}[tb]
  \centering
  \includegraphics[width=1.0\linewidth]{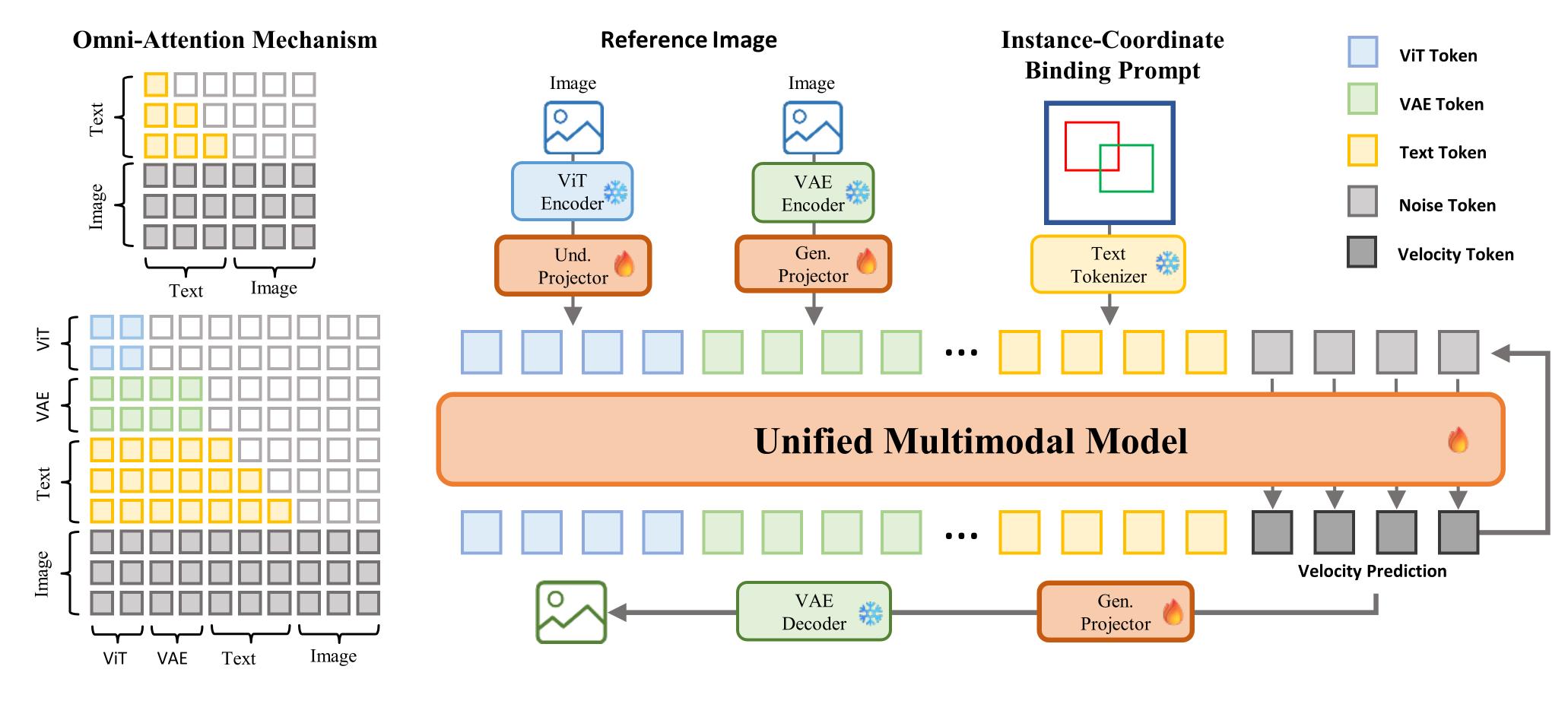}
  \vspace{-8.5mm}
  \caption{Overview of the proposed ConsistCompose framework.}
  \label{fig:framework}
  \vspace{-2.5mm}
\end{figure*}

We present ConsistCompose, a unified multimodal generation framework built upon the MoT architecture of Bagel~\cite{deng2025emerging}. 
At its core lies the proposed LELG paradigm, which enables spatially controllable multi-instance generation by embedding explicit layout semantics—encoded as textual tokens—directly into the linguistic stream. 
This design allows layout control to be natively integrated into a unified multimodal modeling pipeline without introducing task-specific or layout-centric branches. 
An overview of the framework is shown in \cref{fig:framework}.

\subsection{Model Architecture}
ConsistCompose adopts Bagel as its backbone, featuring two Transformer experts within the MoT architecture: one for multimodal understanding and the other for multimodal generation. 
Two visual encoders are employed: a SigLIP-initialized ViT~\cite{tschannen2025siglip} for semantic perception and a FLUX-initialized VAE~\cite{labs2025flux} for image generation. 
Both experts operate on a shared token sequence through unified self-attention, ensuring consistent latent representations across perception and generation. 
All modalities—including text, ViT features, and VAE latents—are projected into a common embedding space for seamless multimodal fusion:
\begin{equation}
\mathbf{Z} = \mathrm{Concat}(\mathbf{T}, \mathbf{V}_{\text{ViT}}, \mathbf{V}_{\text{VAE}}) \in \mathbb{R}^{N\times d},
\label{eq:feature_concat}
\end{equation}
where $\mathbf{T}$ denotes textual embeddings, and $\mathbf{V}_{\text{ViT}}$ and $\mathbf{V}_{\text{VAE}}$ represent visual features for perception and generation, respectively.

\subsection{Linguistic-Embedded Layout-Grounded Generation}
\label{sec:lelg}

The central innovation of ConsistCompose is the LELG paradigm, which achieves spatially controllable generation entirely within the language interface. 
Instead of introducing separate layout encoders or task-specific geometric branches, ConsistCompose embeds normalized bounding boxes directly as  textual tokens.
Each coordinate is represented with three significant decimal places, effectively discretizing the continuous layout space while maintaining strong compatibility with natural language. 
This simple yet powerful formulation unifies spatial reasoning and linguistic grounding within a single generative interface.

Formally, for each instance $i$ with bounding box $b_i=(x_1^i,y_1^i,x_2^i,y_2^i)\in[0,1]^4$, coordinates are inserted immediately after the corresponding subject phrase $s_i$:
\begin{center}
\ttfamily\small
``a brown sofa <bbox>[$x_1^i, y_1^i, x_2^i, y_2^i$]</bbox>''
\end{center}
This text-embedded spatial representation forms the Instance Coordinate Binding Prompt (ICBP), a minimal yet effective mechanism that binds each object's linguistic reference to its spatial coordinates within a unified generative sequence.

During multimodal pretraining, textual and visual tokens are jointly modeled within a unified Transformer sequence:
\begin{equation}
\mathbf{Z} = [\mathbf{T}_{\text{ICBP}}; \mathbf{V}_{\text{ViT}}; \mathbf{V}_{\text{VAE}}],
\label{eq:icbp_sequence}
\end{equation}
where $\mathbf{T}_{\text{ICBP}}$ includes both semantic and coordinate tokens. 
Without additional geometric losses, the Transformer learns spatial grounding implicitly through self-attention over interleaved textual and vision tokens. 
Conditioned on $\mathbf{T}_{\text{ICBP}}$, the generation process is defined as
\begin{equation}
\mathbf{v}_\theta(\mathbf{x}_t \mid \mathbf{T}_{\text{ICBP}}, \mathbf{V}_{\text{ViT}}, \mathbf{V}_{\text{VAE}}),
\label{eq:icbp_v_prediction}
\end{equation}
enabling ConsistCompose to produce multi-instance scenes with spatial precision and semantic consistency, while preserving the simplicity of unified vision-language modeling.

\subsection{Layout Control Enhancement}
We introduce a Coordinate-CFG mechanism to enhance spatial controllability by coupling linguistic and geometric cues. 
Analogous to text- and image-guided CFG, the coordinate-conditioned branch uses prompts containing layout tokens, while the unconditioned branch omits them. 
Let \( \mathbf{v}_{t}^{\text{coord}} \) and \( \mathbf{v}_{t}^{\text{coord-uncond}} \) denote the predicted velocities with and without coordinate conditioning, respectively. 
Coordinate-CFG is formulated as
\begin{equation}
\mathbf{v}_{t}^{\text{coord-cfg}} 
= \mathbf{v}_{t}^{\text{coord-uncond}} 
+ s_{\text{coord}} \big(\mathbf{v}_{t}^{\text{coord}} - \mathbf{v}_{t}^{\text{coord-uncond}}\big),
\label{eq:coord_cfg}
\end{equation}
where \( s_{\text{coord}} \) controls the strength of spatial guidance: larger values enforce stricter adherence to the specified layout, while smaller values allow more flexible compositions.

To stabilize the guidance magnitude, we further normalize the predicted velocity:
\begin{equation}
\mathbf{v}_{t}^{\text{coord-norm}} 
= \alpha \, \mathbf{v}_{t}^{\text{coord-cfg}}, \quad 
\alpha = 
\frac{\|\mathbf{v}_{t}^{\text{base}}\|_{\mathcal{N}}}
{\|\mathbf{v}_{t}^{\text{coord-cfg}}\|_{\mathcal{N}} + \epsilon},
\label{eq:coord_norm}
\end{equation}
where \( \mathbf{v}_{t}^{\text{base}} \) denotes the velocity predicted by the standard text-guided model, \( \|\cdot\|_{\mathcal{N}} \) is the normalization domain, and \( \epsilon \) is a small constant for numerical stability.

\begin{figure}[tb]
    \centering
    \includegraphics[width=0.96\linewidth]{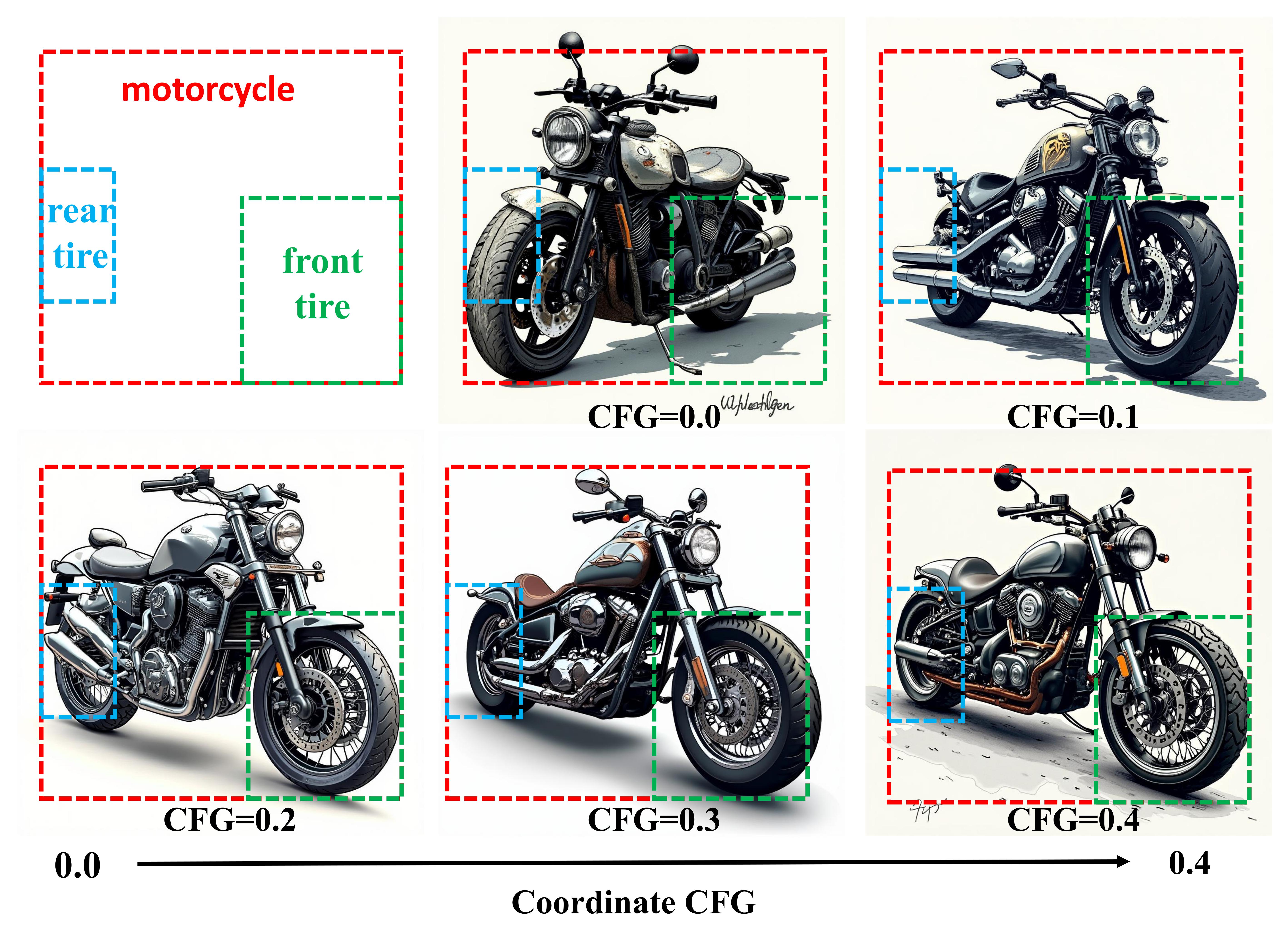}
    \caption{Effect of Coordinate-CFG: larger scale yields progressively more accurate layout alignment}
    \label{fig:coord_cfg}
\end{figure}

\paragraph{Effect of Coordinate-CFG.}
As illustrated in Fig.~\ref{fig:coord_cfg}, increasing \( s_{\text{coord}} \) progressively improves positional accuracy and layout fidelity until overly strong guidance begins to slightly degrade perceptual quality. 
This confirms that Coordinate-CFG effectively injects explicit layout cues into the unified vision-language interface, enabling precise and stable layout-controlled generation.

\subsection{Training Objectives}
Training combines Flow Matching for image generation and autoregressive token prediction for multimodal understanding.
Given a text-image pair $(\mathbf{c}, \mathbf{x})$, Flow Matching~\cite{lipman2022flow} interpolates between latent states:
\begin{equation}
\mathbf{x}_t = (1 - t)\mathbf{x}_0 + t\mathbf{x}_1, \quad t \sim \mathcal{U}(0,1),
\label{eq:flow_matching}
\end{equation}
where $\mathbf{x}_0$ and $\mathbf{x}_1$ denote source and target latents, and the model predicts the corresponding velocity field:
\begin{equation}
\mathcal{L}_{\text{FM}} = 
\mathbb{E}_{t,\mathbf{x}_0,\mathbf{x}_1}\!
\left[\big\|\mathbf{v}_\theta(\mathbf{x}_t \mid \mathbf{c}) - (\mathbf{x}_0 - \mathbf{x}_1)\big\|_2^2\right].
\label{eq:flow_matching_loss}
\end{equation}
We follow the standard formulation of Flow Matching~\cite{lipman2022flow} without architectural modifications.

For multimodal understanding, ConsistCompose performs next-token prediction over the interleaved sequence of text and image tokens:
\begin{equation}
\mathcal{L}_{\text{LM}} = -\!\!\sum_{i=1}^{N}\log P_\theta(z_i \mid z_{<i}),
\label{eq:lm_loss}
\end{equation}
where $\{z_i\}$ denotes the mixed token sequence.  
The total objective is a weighted combination:
\begin{equation}
\mathcal{L} = \lambda_{\text{FM}}\mathcal{L}_{\text{FM}} + \lambda_{\text{LM}}\mathcal{L}_{\text{LM}}.
\label{eq:total_loss}
\end{equation}
No additional coordinate regression loss is introduced, and the model learns spatial grounding purely from layout cues injected into the language stream. 

\subsection{Training Strategy}
We employ a two-stage training strategy.  
In the \emph{alignment} stage, we train on a mixture of general multimodal understanding data to retain broad reasoning ability and ConsistCompose3M to inject layout awareness into both the understanding and generation experts.  
In the subsequent \emph{hybrid supervised fine-tuning} (SFT) stage, we jointly train on general understanding, generation, editing, and multi-subject reference generation data, together with ConsistCompose3M.  
High-resolution fine-tuning is further performed to balance performance across layout control and general-purpose image generation.  
Detailed training configurations are provided in the appendix.
\begin{figure}[tb]
    \centering
    \includegraphics[width=\linewidth]{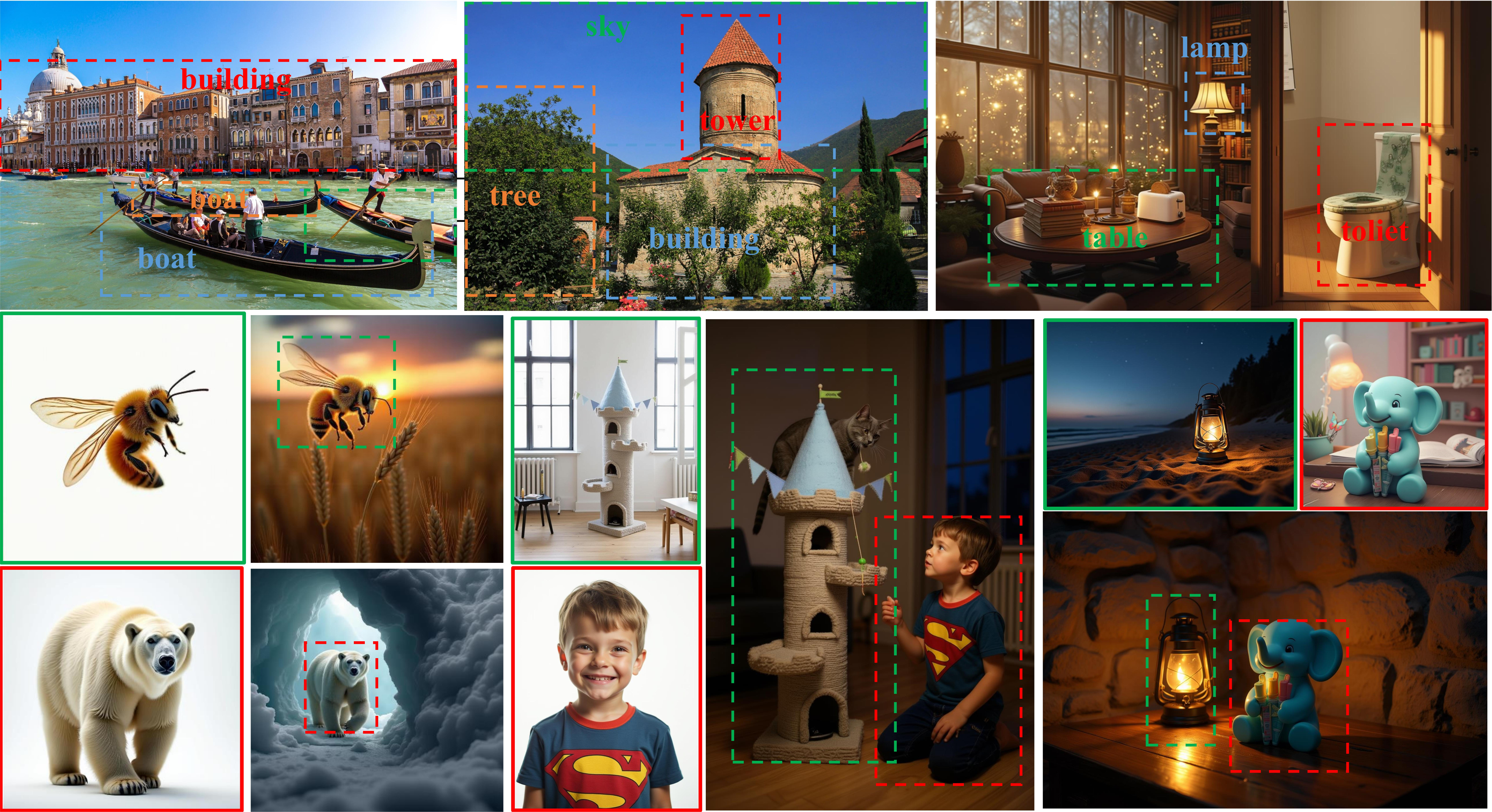}
    \caption{ConsistCompose3M data gallery: examples of layout-grounded text-to-image (T2I) generation and multi-reference image-guided composition.}
    \label{fig:data_gallery}
\end{figure}
\subsection{ConsistCompose3M Dataset} 
We introduce ConsistCompose3M, a large-scale dataset for layout-controllable multi-instance image generation that advances prior work in terms of \emph{scale}, \emph{quality}, and \emph{adaptability}: it contains millions of diverse multi-instance scenes, provides identity-preserving samples filtered by similarity, offers structured spatial and semantic supervision suitable for unified multimodal training, and consists of two complementary splits—a layout-grounded text-to-image split and a reference-conditioned split for subject-preserving layout-guided generation—with representative samples showcased in the data gallery (Figure \ref{fig:data_gallery}).

\begin{table*}[!t]
  \centering
  \footnotesize
  \setlength{\tabcolsep}{3pt}
  \renewcommand{\arraystretch}{1.0}
  \caption{Layout control performance on COCO-Position. 
  We report Instance and Image Success Ratios, as well as Position Accuracy (mIoU, AP, AP50, AP75). 
  \best{Best} results are in bold and \second{second-best} are underlined.}
  \label{tab:coco_position}
  \vspace{0mm}
  \begin{tabularx}{\linewidth}{
    l |
    *{6}{>{\centering\arraybackslash}X} |
    *{6}{>{\centering\arraybackslash}X} |
    *{4}{>{\centering\arraybackslash}X}
  }
    \toprule
    \multicolumn{1}{c|}{\textbf{Methods}} &
    \multicolumn{6}{c|}{\textbf{Instance Success Ratio} (\%) $\uparrow$} &
    \multicolumn{6}{c|}{\textbf{Image Success Ratio} (\%) $\uparrow$} &
    \multicolumn{4}{c}{\textbf{Position Accuracy} (\%) $\uparrow$} \\
    \cmidrule(lr){1-1}\cmidrule(lr){2-7}\cmidrule(lr){8-13}\cmidrule(lr){14-17}
     & \mbox{$L_2$} & \mbox{$L_3$} & \mbox{$L_4$} & \mbox{$L_5$} & \mbox{$L_6$} & \mbox{\textbf{Avg}}
     & \mbox{$L_2$} & \mbox{$L_3$} & \mbox{$L_4$} & \mbox{$L_5$} & \mbox{$L_6$} & \mbox{\textbf{Avg}}
     & \mbox{mIoU} & \mbox{AP} & \mbox{AP50} & \mbox{AP75} \\
    \midrule
    GLIGEN
    & 89.1 & 86.3 & 82.0 & 79.6 & 81.6 & 82.6
    & 78.8 & 63.8 & 48.1 & 35.0 & 35.0 & 52.1
    & 69.0 & 40.5 & 75.9 & 39.1 \\
    InstanceDiffusion
    & \second{94.1} & \best{94.4} & \second{89.5} & \second{84.6} & \second{83.8} & \second{87.8}
    & \second{89.4} & \best{84.4} & \second{67.5} & 46.9 & \second{39.4} & \second{65.5}
    & \second{78.1} & \second{57.2} & \second{83.6} & \second{65.5} \\
    MIGC++
    & \second{94.1} & 92.1 & 87.3 & 84.1 & 83.4 & 86.8
    & \second{89.4} & 78.1 & 62.5 & \second{48.1} & 38.8 & 63.4
    & 74.9 & 48.3 & 79.2 & 52.6 \\
    CreatiLayout
    & 81.9 & 76.3 & 73.4 & 73.5 & 71.2 & 74.0
    & 69.4 & 48.1 & 36.9 & 31.9 & 26.3 & 42.5
    & 64.9 & 32.4 & 61.1 & 31.6 \\
    PlanGen
    & 85.3 & 84.2 & 83.8 & 80.9 & 81.2 & 82.5
    & 72.5 & 63.1 & 51.3 & 33.1 & 31.3 & 50.3
    & 66.2 & 31.9 & 74.0 & 21.5 \\
    \midrule
    Ours
    & \best{95.6} & \second{94.2} & \best{92.7} & \best{90.6} & \best{92.4} & \best{92.6}
    & \best{91.9} & \second{83.1} & \best{73.1} & \best{63.7} & \best{68.8} & \best{76.1}
    & \best{85.3} & \best{70.9} & \best{89.1} & \best{76.9} \\
    \bottomrule
  \end{tabularx}
  \vspace{-3mm}
\end{table*}

\begin{figure*}[!t]
  \centering
  \footnotesize
  \setlength{\tabcolsep}{0.5pt}
  \renewcommand{\arraystretch}{0.9}
  \begin{tabular}{@{}cccccccc@{}}
    \makecell[c]{\tiny\textbf{Annotation}} &
    \makecell[c]{\tiny\textbf{GLIGEN}} &
    \makecell[c]{\tiny\textbf{InstanceDiffusion}} &
    \makecell[c]{\tiny\textbf{MIGC++}} &
    \makecell[c]{\tiny\textbf{CreatiLayout}} &
    \makecell[c]{\tiny\textbf{PlanGen}} &
    \makecell[c]{\tiny\textbf{Alignment}} &
    \makecell[c]{\tiny\textbf{Hybrid SFT}} \\[0.8mm]

    \includegraphics[width=0.123\textwidth]{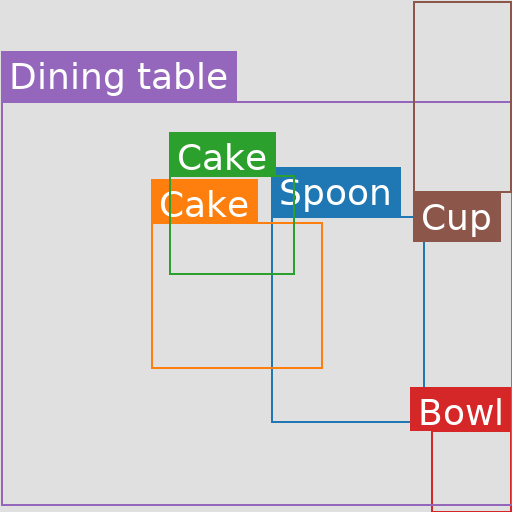} &
    \includegraphics[width=0.123\textwidth]{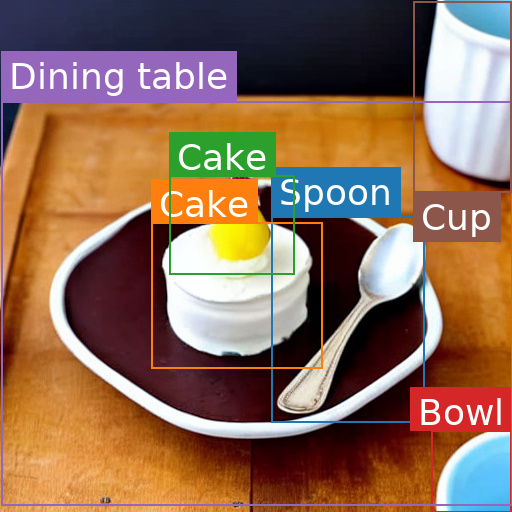} &
    \includegraphics[width=0.123\textwidth]{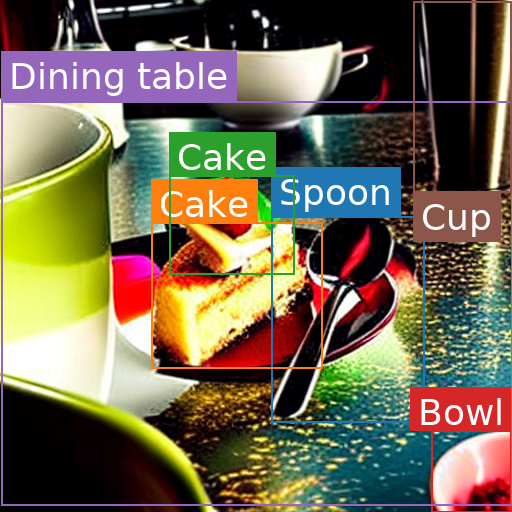} &
    \includegraphics[width=0.123\textwidth]{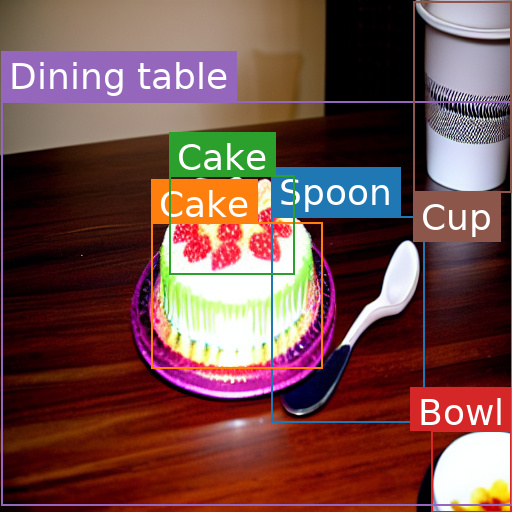} &
    \includegraphics[width=0.123\textwidth]{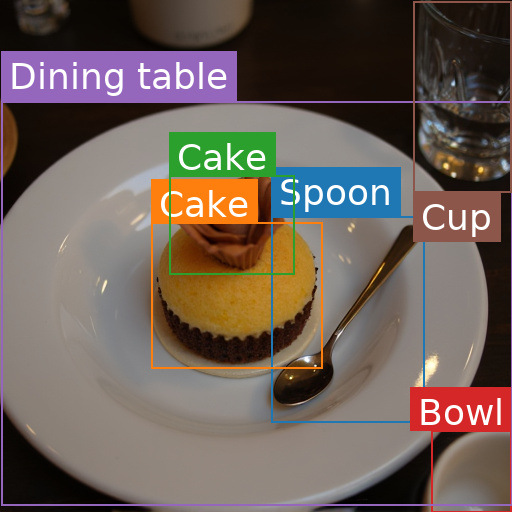} &
    \includegraphics[width=0.123\textwidth]{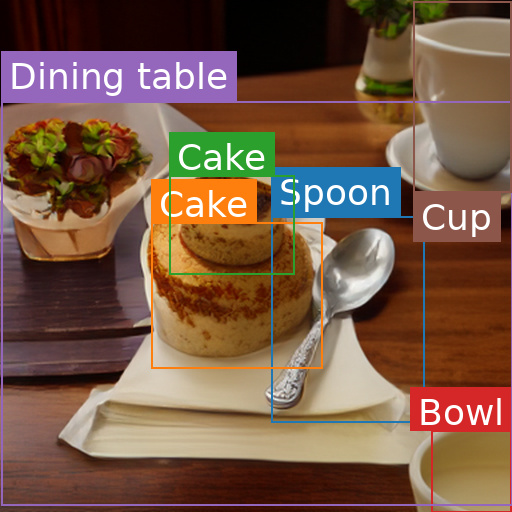} &
    \includegraphics[width=0.123\textwidth]{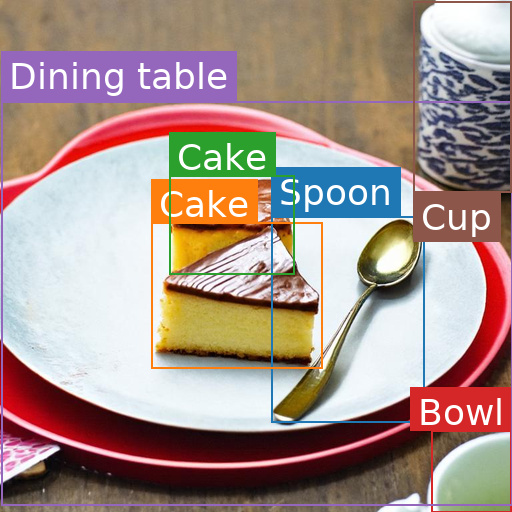} &
    \includegraphics[width=0.123\textwidth]{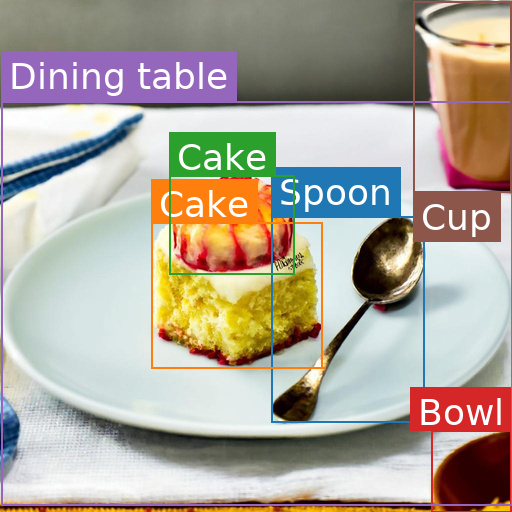} \\[-0.5mm]

    \includegraphics[width=0.123\textwidth]{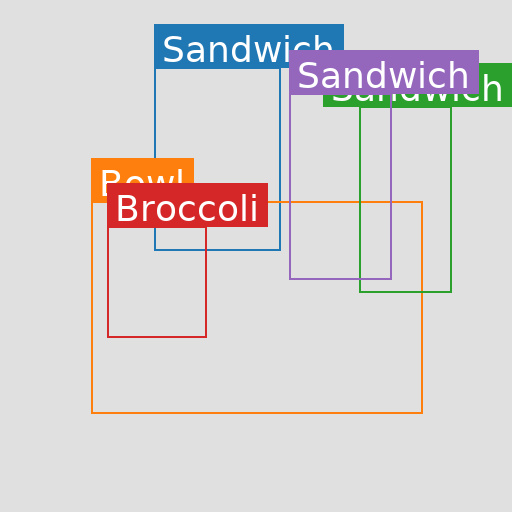} &
    \includegraphics[width=0.123\textwidth]{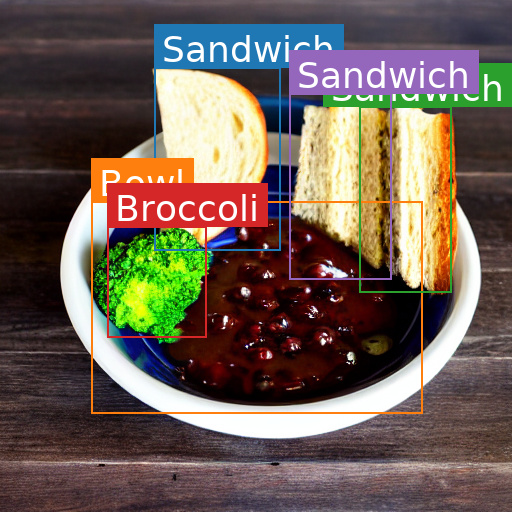} &
    \includegraphics[width=0.123\textwidth]{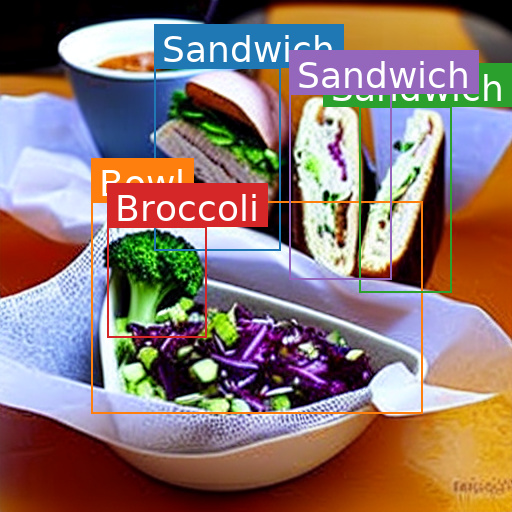} &
    \includegraphics[width=0.123\textwidth]{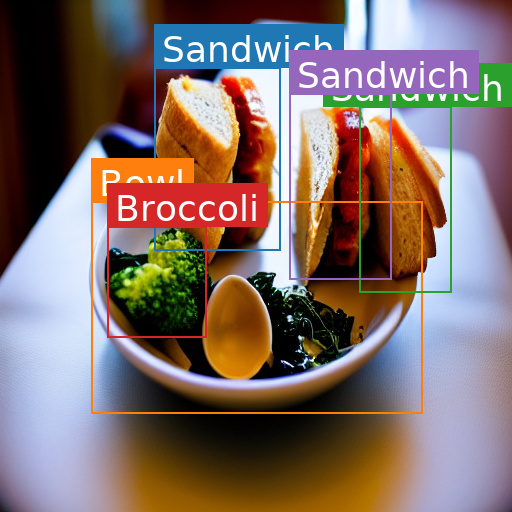} &
    \includegraphics[width=0.123\textwidth]{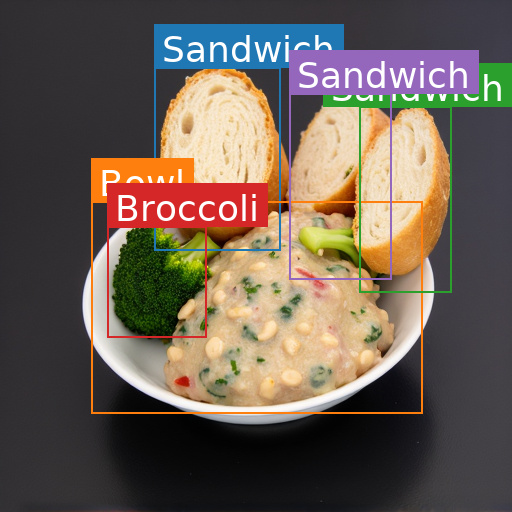} &
    \includegraphics[width=0.123\textwidth]{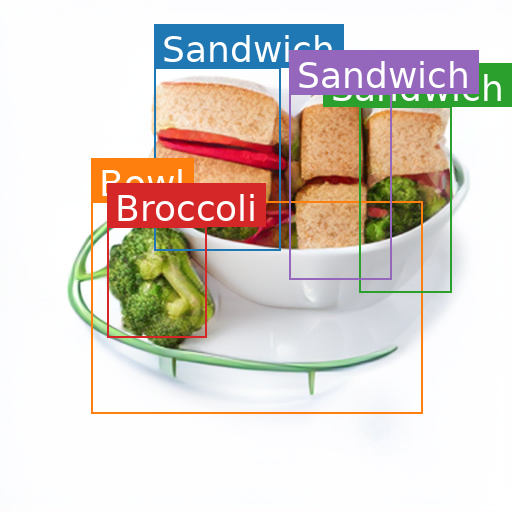} &
    \includegraphics[width=0.123\textwidth]{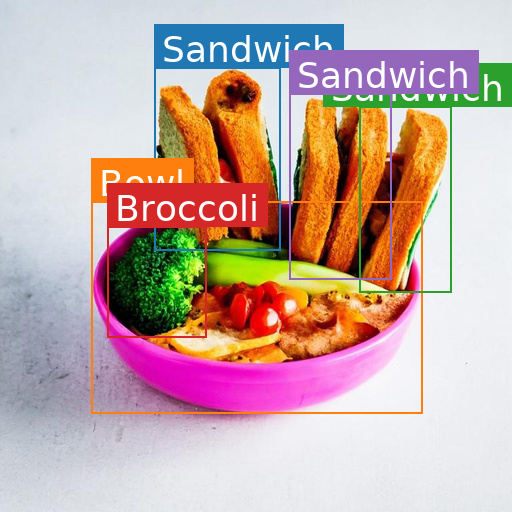} &
    \includegraphics[width=0.123\textwidth]{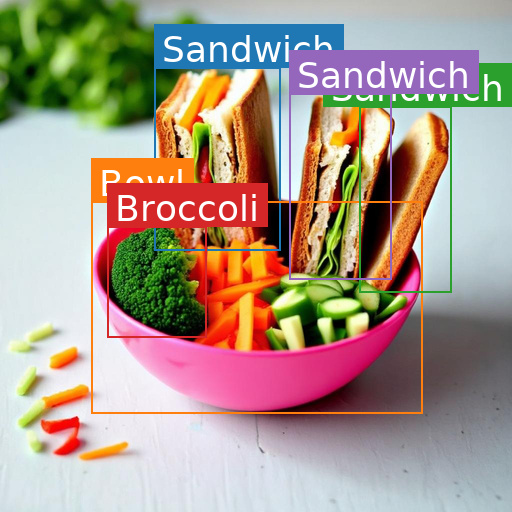} \\[-0.5mm]

    \includegraphics[width=0.123\textwidth]{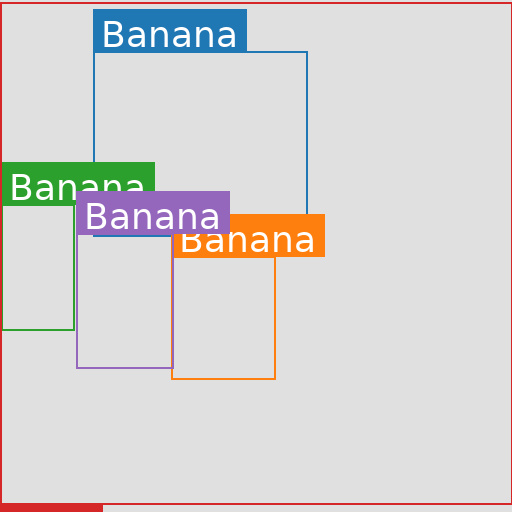} &
    \includegraphics[width=0.123\textwidth]{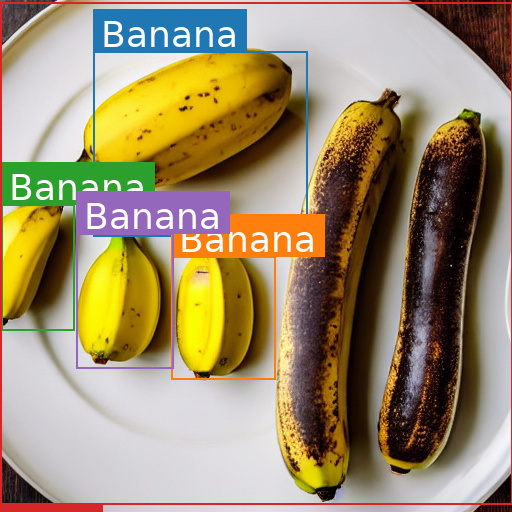} &
    \includegraphics[width=0.123\textwidth]{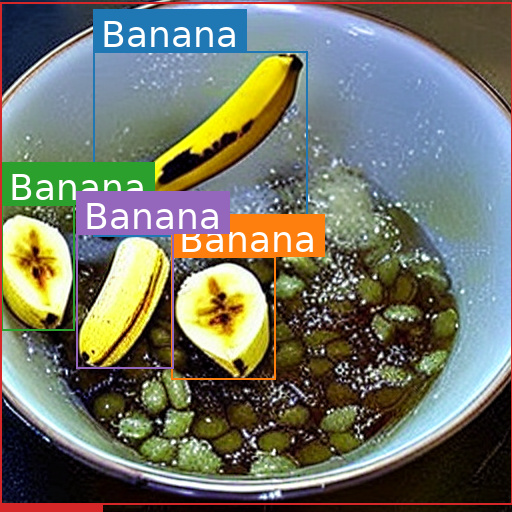} &
    \includegraphics[width=0.123\textwidth]{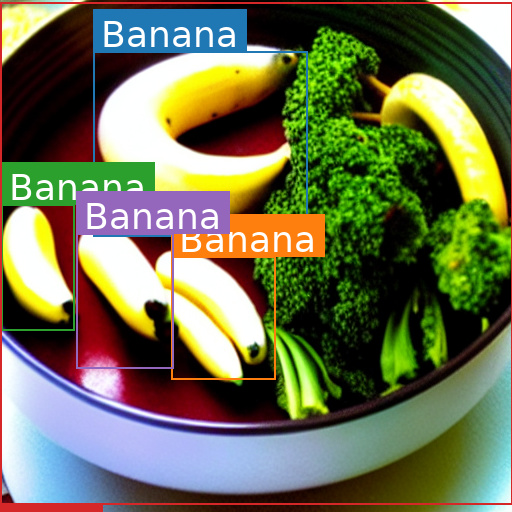} &
    \includegraphics[width=0.123\textwidth]{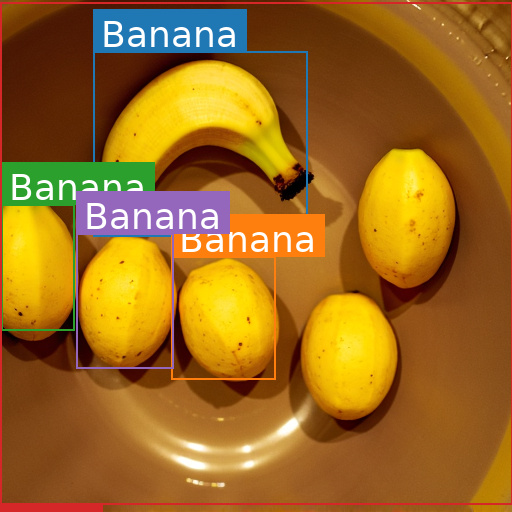} &
    \includegraphics[width=0.123\textwidth]{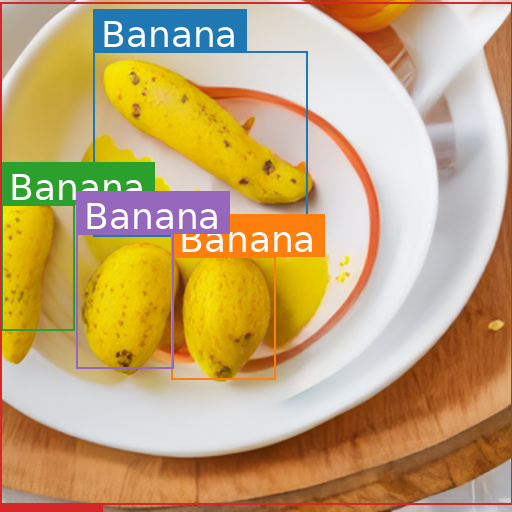} &
    \includegraphics[width=0.123\textwidth]{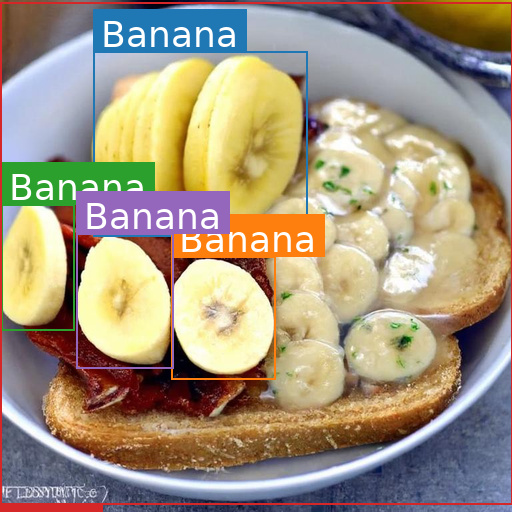} &
    \includegraphics[width=0.123\textwidth]{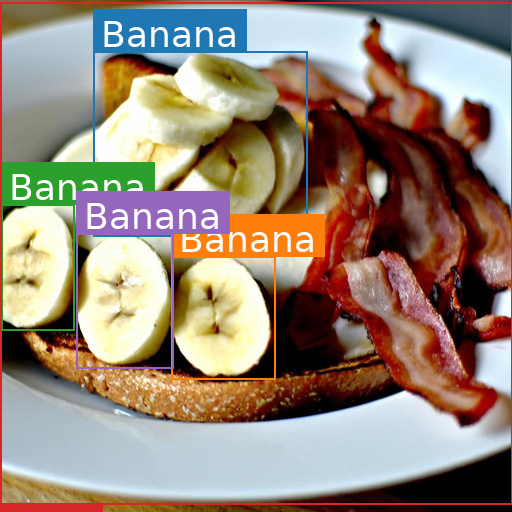}
  \end{tabular}
  \vspace{-1mm}
  \caption{Qualitative Comparison on COCO-Position: ConsistCompose respects target layouts with coherent and detailed scenes.}
  \label{fig:coco_position}
  \vspace{-1mm}
\end{figure*}

The T2I component is obtained by reprocessing LayoutSAM~\cite{zhang2025creatilayout}, which provides diverse multi-object layouts with instance-level annotations.
For each instance, we append its normalized bounding box to the caption via the ICBP mechanism (Sec.~\ref{sec:lelg}), injecting layout coordinates into the prompt and turning dense layout annotations into language-embedded supervision without extra coordinate channels or layout-specific encoders.
The reference-conditioned split targets identity-preserving layout control. We reuse subject-centric assets from existing pipelines—subject banks and crops from Subjects200K~\cite{tan2025ominicontrol} and UNO~\cite{wu2025less}, defined on the Objects365 taxonomy—rather than generating new subjects.
These subjects are recomposed into multi-subject scenes under diverse layouts, with bounding boxes and explicit pairings between each scene and its reference images. We then attach ICBP-style captions and filter samples with CLIP/DINO-based similarity to enforce identity consistency and discard layout-inconsistent or low-quality examples.
Taken together, these two components constitute a large-scale corpus with tightly aligned spatial and semantic supervision, supporting both text-only and reference-conditioned layout-aware generation and providing a principled testbed for studying layout-controllable generation in unified multimodal models.

\section{Experiment}
\label{sec:experiment}

We conduct extensive experiments to evaluate ConsistCompose on both layout-controlled generation and general multimodal understanding. We benchmark on COCO-Position~\cite{lin2014microsoft, Zhou2024MIGCMG} and MS-Bench~\cite{wang2025msdiffusion} for text- and vision-conditioned multi-instance layout control, and assess general capabilities on GenEval~\cite{ghosh2023geneval}, GEdit~\cite{liu2025step1x}, MMMU~\cite{yue2024mmmu}, MMBench~\cite{liu2024mmbench}, and DreamBench~\cite{ruiz2023dreambooth}. Together, these evaluations test (i) fine-grained spatial control, (ii) identity-preserving multi-subject generation, and (iii) retention of broad reasoning ability under unified multimodal training.

\subsection{Implementation Details}
\begin{table*}[!tb]
  \centering
  \footnotesize
  \setlength{\tabcolsep}{6pt}
  \renewcommand{\arraystretch}{1.12}
  \caption{Performance on MS-Bench and MS-Bench-Random. Best values are \best{bold}, and second-best values are \second{underlined}.}
  \label{tab:msbench}
  \begin{tabularx}{\linewidth}{
    l |
    *{4}{>{\centering\arraybackslash}X} |
    *{4}{>{\centering\arraybackslash}X}
  }
    \toprule
    \multicolumn{1}{c|}{\textbf{Methods}} &
    \multicolumn{4}{c|}{\textbf{MS-Bench}} &
    \multicolumn{4}{c}{\textbf{MS-Bench-Random}} \\
    \cmidrule(lr){2-5}\cmidrule(lr){6-9}
     & \mbox{\textbf{CLIP-T}} & \mbox{\textbf{DINO}} & \mbox{\textbf{mIoU}} & \mbox{\textbf{AP}}
     & \mbox{\textbf{CLIP-T}} & \mbox{\textbf{DINO}} & \mbox{\textbf{mIoU}} & \mbox{\textbf{AP}} \\
    \midrule
    GLIGEN           & 0.309 & 0.454 & \second{0.868} & \second{0.751}
                     & 0.312 & 0.431 & \second{0.858} & \second{0.722} \\
    MS-Diffusion     & \best{0.336} & 0.555 & 0.466 & 0.108
                     & \best{0.334} & 0.544 & 0.464 & 0.105 \\
    MUSE             & 0.320 & \second{0.619} & 0.698 & 0.352
                     & 0.321 & \second{0.607} & 0.673 & 0.303 \\
    \midrule
    Ours & \second{0.333} & \best{0.660} & \best{0.889} & \best{0.789} 
         & \best{0.334} & \best{0.630} & \best{0.878} & \best{0.756} \\
    \bottomrule
  \end{tabularx}
\end{table*}
\noindent\textbf{Training Data.} 
We initialize our model from the Bagel foundation model and primarily train it on ConsistCompose3M. 
To preserve broad visual–linguistic understanding, we additionally incorporate FineVision~\cite{Wiedmann2025FineVisionOD} and MAmmoTH~\cite{yue2023mammoth}. 
To better support multi-instance layout reasoning, we further mine high-resolution ($>512$ px) images from COCO2017~\cite{lin2015microsoftcococommonobjects} and Objects365~\cite{shao2019objects365}, discarding low-quality samples. 
These corpora provide dense multi-object scenes and same-category groupings that strengthen the model’s understanding of parallel layouts within the LELG framework.

In the second-stage hybrid SFT, we interleave two complementary datasets: GPT-Image-Edit-1.5M~\cite{wang2025gptimageedit15mmillionscalegptgeneratedimage} to retain editing capabilities and Text-to-Image-2M~\cite{jackyhate_text-to-image-2M_2024} to further enhance text-driven generation. This two-stage schedule allows us to inject layout-awareness while preserving the strong general abilities of the base model.

\noindent\textbf{Training and Inference Configurations.} 
Both alignment and SFT stages use AdamW with $\beta_1=0.9$, $\beta_2=0.95$, $\epsilon=1.0\times10^{-15}$, and a fixed learning rate of $2.5\times10^{-5}$. Training is performed on 64 GPUs with sequence packing to balance throughput and memory. At inference time, we apply Coordinate-CFG (Sec.~\ref{sec:coord_cfg_ablation}) to strengthen layout adherence, setting the CFG scale to 1.6 for COCO-Position and 0.6 for MS-Bench. Full configuration details are provided in the supplementary material.

\subsection{Evaluation Metrics and Benchmarks}

\noindent\textbf{COCO-Position.} 
Spatial accuracy on COCO-Position is evaluated using \emph{Instance Success Ratio}, \emph{Image Success Ratio}, and \emph{Position Accuracy}. Instance Success Ratio is computed over instances whose predicted bounding boxes achieve $\text{IoU}>0.5$ (via Grounding DINO~\cite{liu2024grounding}) and is reported across different instance levels $L_2$--$L_6$. Image Success Ratio additionally requires all instances in an image to be successfully localized, capturing global layout coherence. Position Accuracy is measured by mIoU and AP, reflecting fine-grained alignment between predicted and target box locations across varying IoU thresholds.

\noindent\textbf{MS-Bench.} 
MS-Bench~\cite{peng2025muse, wang2025msdiffusion} contains 1{,}148 multi-subject combinations (2-3 subjects per image) from 40 categories, each with ground-truth bounding boxes. We follow prior work and evaluate three aspects: (1) \emph{positional accuracy}, (2) \emph{text-image consistency}, and (3) \emph{identity consistency}. CLIP-T~\cite{hessel2022clipscorereferencefreeevaluationmetric} measures global text-image semantic alignment, DINOv2~\cite{oquab2023dinov2} captures fine-grained correspondence between cropped generations and reference subjects, and IoU/AP (via Grounding DINO) quantify localization accuracy under multi-subject layouts. We report results on both the original MS-Bench split and the MS-Bench-Random variant, which shuffles object layouts to stress-test spatial generalization while keeping subject identities fixed.

\begin{figure}[!htbp]
  \centering
  \includegraphics[width=1.0\linewidth]{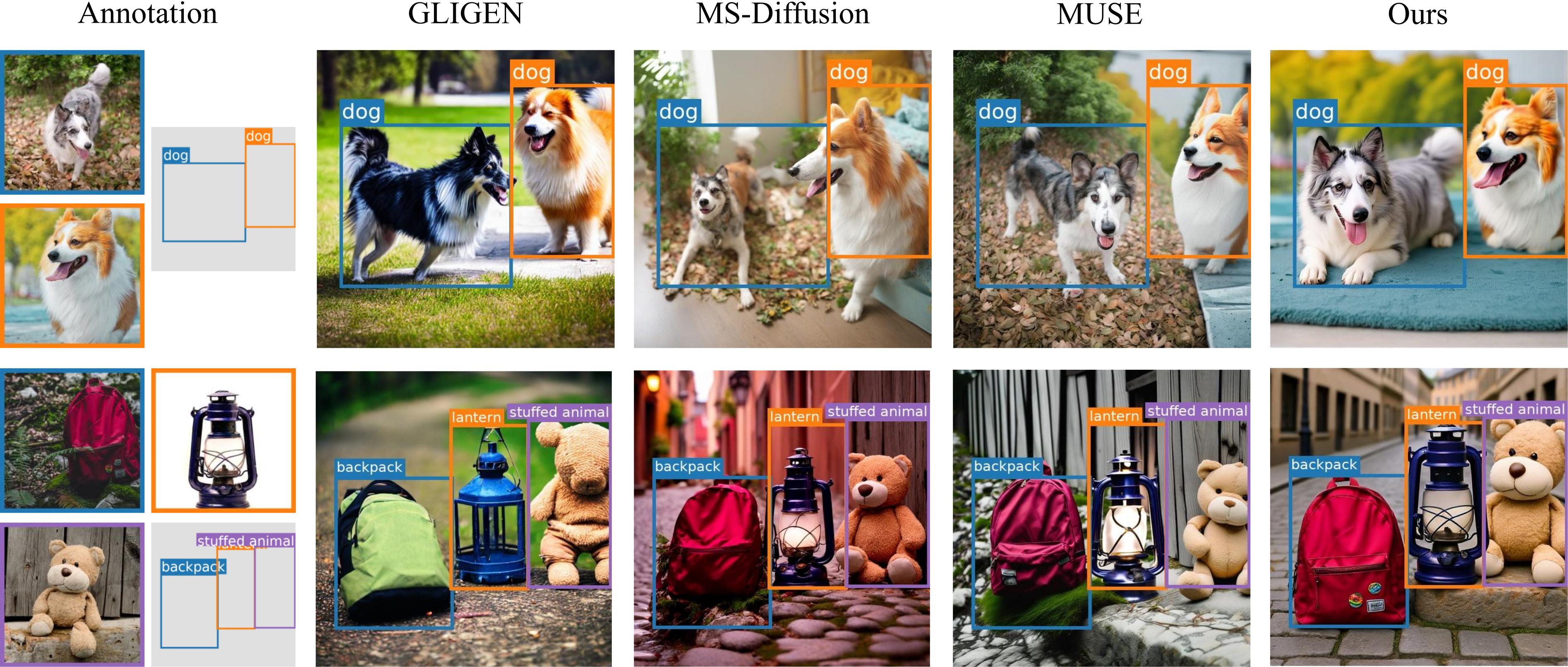}
  \caption{Qualitative comparison on MS-Bench-Random. Our model preserves reference identity while faithfully respecting randomized layouts.}
  \label{fig:msbench}
\end{figure}

\noindent\textbf{General Capabilities.} 
To verify that layout-aware training does not degrade broad multimodal competence, we evaluate on MMMU, MMBench, GenEval, GEdit, and DreamBench~\cite{yue2024mmmu, liu2024mmbench, ghosh2023geneval,liu2025step1x,ruiz2023dreambooth}. MMMU is reported with Task Accuracy to reflect multi-domain reasoning; MMBench uses the Overall Comprehension Score to evaluate visual understanding and cross-modal grounding; GenEval provides object-centric metrics for counting, spatial relations, attribute binding, and negation; GEdit evaluates instruction-driven image editing quality; 
DreamBench assesses subject-driven generation under single- and multi-object settings.

\subsection{Results on Layout-Controlled Generation}

We compare ConsistCompose with leading layout-aware and multi-subject generation models, including GLIGEN, InstanceDiffusion, MIGC++, CreatiLayout, PlanGen for layout control, MS-Diffusion and MUSE for multi-instance generation. All baselines are reproduced using official implementations and default configurations for fairness.

\noindent\textbf{COCO-Position.}
As shown in \cref{tab:coco_position}, ConsistCompose achieves the best performance on all COCO-Position metrics. The average \emph{Instance Success Ratio} reaches 92.6\%, surpassing the second-best model by 4.8\%. At the image level, the \emph{Image Success Ratio} attains 76.1\%, indicating substantially stronger global spatial coherence across multiple instances. 

For Position Accuracy, our method yields mIoU of 85.3, AP of 70.9, consistently outperforming all baselines. The gains are particularly pronounced at higher instance counts ($L_4$--$L_6$), where cluttered layouts challenge prior methods. Qualitative examples in \cref{fig:coco_position} illustrate that ConsistCompose produces sharper object boundaries, more faithful spatial arrangements, and more consistent lighting and texture than GLIGEN, InstanceDiffusion, and MIGC++, while the hybrid SFT stage further improves visual fidelity.

\noindent\textbf{MS-Bench and MS-Bench-Random.}
\cref{tab:msbench} summarizes results on MS-Bench and MS-Bench-Random. The baselines exhibit complementary strengths: GLIGEN attains strong mIoU (0.868/0.858) but relatively low DINO (0.454/0.431), indicating accurate localization but weaker identity preservation; MUSE improves DINO (0.619/0.607) with only moderate mIoU (0.698/0.673); MS-Diffusion achieves the highest CLIP-T (0.336/0.334) but performs poorly on spatial metrics.

In contrast, ConsistCompose delivers consistent state-of-the-art performance on both benchmarks. On MS-Bench, it reaches DINO 0.660, mIoU 0.889, and AP 0.789; on MS-Bench-Random, it attains DINO 0.630, mIoU 0.878, and AP 0.756. We lead on all core layout- and identity-related metrics (DINO, mIoU, AP) under both fixed and randomized layouts, showing that LELG and our multi-reference prompting jointly enforce spatial fidelity and subject consistency. Qualitative examples in \cref{fig:msbench} further illustrate that our generations preserve fine-grained appearance details while closely following target layouts, even under large random perturbations.

\subsection{General Multimodal Capability}

\begin{table}[tb]
  \centering
  \small
  \caption{General capability and DreamBench evaluation.}
  \label{tab:general_and_dreambench}
  \setlength{\tabcolsep}{2pt} 
  \renewcommand{\arraystretch}{1.08}
  
  \vspace{2pt}
  \textbf{(a) General capability}\\[2pt]
  \begin{tabular}{l c c c c}
    \toprule
    \textbf{Model} & \textbf{MMBench$\uparrow$} & \textbf{MMMU$\uparrow$} & \textbf{GenEval$\uparrow$} & \textbf{GEdit$\uparrow$}\\
    \midrule
    Bagel Base                & 81.4 & 46.4 & 0.86 & 6.68 \\
    \rowcolor[HTML]{E6F2FF}
    Ours (w/o Coord)     & 81.5 & 39.4 & 0.88 & 6.23 \\
    \rowcolor[HTML]{E6F2FF}
    Ours (w/ Coord)      & 81.4 & 42.3 & 0.88 & 6.31 \\
    \bottomrule
  \end{tabular}
  
  \vspace{6pt}

  {\fontsize{8pt}{9.6pt}\selectfont
  \setlength{\tabcolsep}{3pt}
  \textbf{(b) DreamBench (Single/Multi)}\\[2pt]
  \begin{tabular}{l c c c c c c}
    \toprule
    \multirow{2}{*}{\textbf{Method}} & \multicolumn{3}{c}{\textbf{Single}} & \multicolumn{3}{c}{\textbf{Multi}} \\
    \cmidrule(lr){2-4} \cmidrule(lr){5-7}
    & \textbf{DINO} & \textbf{CLIP-I} & \textbf{CLIP-T} & \textbf{DINO} & \textbf{CLIP-I} & \textbf{CLIP-T} \\
    \midrule
    UNO            & 0.661 & 0.796 & 0.304 & 0.491 & 0.715 & 0.323 \\
    OmniGen        & 0.554 & 0.746 & 0.322 & 0.441 & 0.692 & 0.341 \\
    OmniGen2       & 0.671 & 0.791 & 0.312 & 0.459 & 0.698 & 0.333 \\
    \rowcolor[HTML]{E6F2FF}
    Ours           & 0.677 & 0.792 & 0.314 & 0.506 & 0.703 & 0.335 \\
    \bottomrule
  \end{tabular}}
\end{table}

\noindent\textbf{General Understanding and DreamBench.}
\cref{tab:general_and_dreambench} summarizes general capability and DreamBench performance. ConsistCompose maintains strong multimodal understanding: MMBench, GenEval and GEdit scores remain comparable to the Bagel Base, indicating that introducing layout supervision does not harm global comprehension. The MMMU scores of our variants stay within a similar range to the base model, showing that cross-domain reasoning remains stable even after LELG training. 

On DreamBench, our model achieves the best DINO and CLIP-I scores in both single- and multi-object settings, indicating stronger instance-level spatial control and identity preservation. These gains are obtained without any DreamBench-specific tuning, arising solely from ConsistCompose3M and unified LELG training.

\subsection{Balance in Coordinate-CFG}
\label{sec:coord_cfg_ablation}

\begin{figure}[tb]
  \centering
  \includegraphics[width=1.0\linewidth]{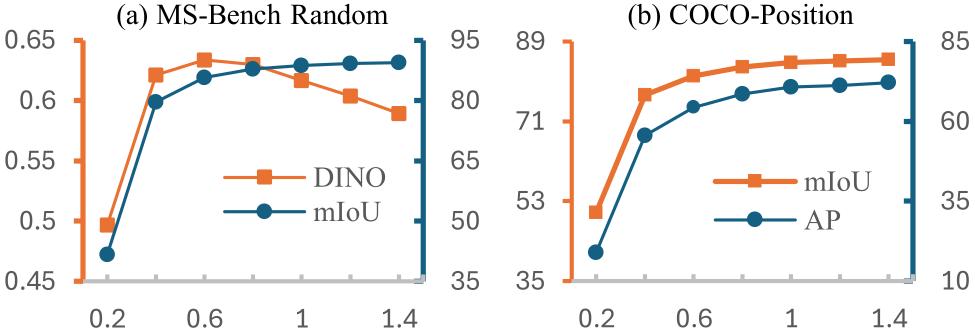}
  \vspace{-4.5mm}
  \caption{Effect of Coordinate-CFG strength. Moderate guidance balances spatial accuracy and identity preservation across COCO-Position and MS-Bench-Random.}
  \label{fig:coord_cfg_ablation}
  \vspace{-2.5mm}
\end{figure}
We investigate how the Coordinate-CFG scale influences two key objectives: spatial placement and image fidelity.

On COCO-Position (text-to-image), increasing the CFG scale from 0.2 to 1.4 steadily improves Position Accuracy, as reflected by mIoU and AP. When the scale exceeds 2.0, however, perceptual quality starts to degrade, revealing a trade-off between strict layout adherence and visual realism. A moderate scale around 1.6 therefore offers the best compromise for T2I generation.

On MS-Bench-Random (multi-reference generation), raising the scale from 0.2 to 0.6 increases the DINO score (identity preservation) from 0.497 to 0.630 and mIoU from 41.7\% to 87.8\%. Further increasing the scale to 1.4 yields only marginal mIoU gains (87.8\% $\to$ 89.4\%) while reducing DINO (0.630 $\to$ 0.589), indicating that overly strong guidance may compromise instance fidelity. These results suggest using a CFG scale in the range 0.4–0.8 for multi-reference generation, which provides accurate layouts while preserving subject identity.
\section{Conclusion}
\label{sec:conclusion}

We presented ConsistCompose, a unified multimodal framework that encodes instance-level layout constraints via interleaved image-text prompts. LELG embeds bounding boxes as textual tokens, enabling spatial and identity control without layout-specific encoders, and is trained with ConsistCompose3M, a 3.4M-sample dataset with structured layout annotations. This interface improves layout-controllable generation, yielding a 7.2\% gain in layout IoU and a 13.7\% AP improvement on COCO-Position while maintaining strong general multimodal performance. Although the current formulation focuses on bounding-box control, LELG provides an extensible abstraction for injecting spatial structure into unified multimodal models, pointing toward finer-grained spatial reasoning, part-aware manipulation, and interactive layout-guided scene composition.
{
    \small
    \bibliographystyle{ieeenat_fullname}
    \bibliography{main}
}

\appendix
\setcounter{page}{0}
\setcounter{section}{0}
\renewcommand{\thefigure}{S\arabic{figure}}
\renewcommand{\thetable}{S\arabic{table}} 

\maketitlesupplementary

\section{Training Recipe of ConsistCompose}
\label{sec:training_recipe_supp}
We adopt a two-stage training recipe for ConsistCompose that injects layout awareness while preserving general multimodal understanding and generation capabilities. Unless otherwise specified, both stages are trained on 64 GPUs with the AdamW optimizer, gradient clipping set to $1.0$, a diffusion timestep shift of $4.0$, and a CE:MSE loss weight of $0.25{:}1$ for the generative objectives. Detailed per-task sampling ratios and stage-specific hyperparameters are summarized in \cref{tab:training_recipe}.

\subsection{Stage 1: LELG Alignment}
\label{sec:stage1_lelg_alignment}

In the first stage, we train on a mixture of general multimodal understanding datasets—such as FineVision~\cite{Wiedmann2025FineVisionOD} and MAmmoTH~\cite{yue2023mammoth}—together with our proposed ConsistCompose3M dataset. This joint training retains broad visual--linguistic reasoning capability while injecting layout awareness into both the understanding and generation experts. 

To further strengthen multi-instance layout reasoning, we additionally mine high-resolution images ($>512$\,px) from COCO2017~\cite{lin2015microsoftcococommonobjects} and Objects365~\cite{shao2019objects365}, discarding low-quality samples. These corpora provide dense multi-object scenes (e.g., street environments with pedestrians, cars, bicycles, and traffic lights) as well as structured same-category groupings such as:
\begin{center}
\ttfamily\small
3 dogs \texttt{<bbox>}$[x_1, y_1, x_2, y_2]$\texttt{</bbox>}, \texttt{<bbox>}$[x_1, y_1, x_2, y_2]$\texttt{</bbox>}, \texttt{<bbox>}$[x_1, y_1, x_2, y_2]$\texttt{</bbox>}
\end{center}
which improve the model's ability to reason about parallel, multi-instance layouts within the LELG framework.

Stage~1 is trained for 18K steps using a cosine learning rate schedule (initial learning rate $2\times10^{-5}$), no weight decay, and AdamW hyperparameters $(\beta_1 = 0.9, \beta_2 = 0.95, \epsilon = 1.0\times10^{-15})$. The sequence length per rank ranges from 30K to 38K tokens. For resolution settings, the generation module adopts $(448, 768)$ as its minimum short side and maximum long side, while the understanding module uses $(224, 386)$.

\subsection{Stage 2: Hybrid Supervised Fine-Tuning}
\label{sec:stage2_sft}

We further train ConsistCompose on a mixture of generation and understanding
tasks, following the hybrid multi-task supervised fine-tuning (SFT) recipe used
in Bagel~\cite{deng2025emerging}. On the generation side, the mixture comprises text-to-image, multi-reference-to-image, LELG-conditioned text-to-image, LELG-conditioned multi-reference-to-image, and image editing, with layout-conditioned tasks upsampled to reinforce layout consistency. On the understanding side, image-to-text pairs and interleaved understanding data are sampled at lower ratios so as to maintain general visual--language competence while keeping the overall training focused on layout-controllable generation.

Stage~2 is trained for 6K steps with a constant learning rate of $2.0\times10^{-5}$, sequence lengths per rank ranging from 36K to 45K tokens, a generation resolution of $(512, 1024)$, and an understanding resolution of $(224, 386)$.

\begin{table}[t]
\centering
\scriptsize
\setlength{\tabcolsep}{3pt} 
\caption{Training recipe of ConsistCompose.}
\label{tab:training_recipe}
\begin{tabular}{p{0.50\linewidth}cc}
\toprule
 & \textbf{LELG Align.} & \textbf{SFT} \\
\midrule
\multicolumn{3}{l}{\textbf{Hyperparameters}} \\
\midrule
Learning rate  & $2\times10^{-5}$ & $2.0\times10^{-5}$ \\
LR scheduler   & Cosine & Constant \\
Weight decay   & 0.0  & 0.0 \\
Gradient norm clip  & 1.0 & 1.0 \\
Optimizer       & \multicolumn{2}{c}{AdamW} \\
Loss weight (CE : MSE)  & -- & 0.25 : 1  \\
Warm-up steps   & 500 & 500 \\
Training steps  & 18K & 6K \\
Seq. length / rank (min, max) & (30K, 38K) & (36K, 45K) \\
Gen res. (short side, long side) & (448, 768) & (512, 1024) \\
Und res. (short side, long side) & (224, 386) & (224, 386) \\
Diffusion timestep shift & 4.0 & 4.0 \\
\midrule
\multicolumn{3}{l}{\textbf{Data sampling ratio}} \\
\midrule
Text-to-Text                       & 0.05 & 0.1 \\
Text-to-Image                      & 0.0 & 0.1 \\
Multi Ref$\rightarrow$Image        & 0.0 & 0.1 \\
Text-to-Image (LELG)               & 0.4 & 0.4 \\
Multi Ref$\rightarrow$Image (LELG) & 0.4 & 0.1 \\
Image Editing                      & 0.0 & 0.1 \\
Image-to-Text pair                 & 0.1 & 0.1 \\
Interleaved understanding          & 0.05 & 0.1 \\
\bottomrule
\end{tabular}
\end{table}

\subsection{Data Sampling}
\label{sec:data_sampling}

Table~\ref{tab:training_recipe} summarizes the per-task sampling ratios and hyperparameters for both stages. During SFT, layout-conditioned tasks are assigned higher sampling ratios to enhance multi-instance layout controllability, whereas general understanding and interleaved tasks are sampled more sparsely to preserve broad multimodal knowledge. 

This two-stage training strategy yields a favorable balance between layout consistency, high-fidelity image generation, and general multimodal understanding.

\section{Details of Coordinate-CFG}
\label{sec:coord_cfg_supp}

\begin{table}[!t]
\small
\centering
\caption{List of symbols and their descriptions.}
\label{tab:symbol_table}
\begin{tabularx}{0.45\textwidth}{@{}lX@{}}
\toprule
Symbol & Description \\
\midrule
$\mathbf{v}_t$ & Predicted velocity with full conditioning (text, image, and coordinates) \\
$\mathbf{v}_t^{\text{text-drop}}$ & Predicted velocity without text and coordinate (image-only branch) \\
$\mathbf{v}_t^{\text{img-drop}}$ & Predicted velocity without image \\
$\mathbf{v}_t^{\text{coord-drop}}$ & Predicted velocity without coordinate \\
$\mathbf{v}_t^{\text{text-cfg}}$ & Text-guided velocity in the hierarchical fusion \\
$\mathbf{v}_t^{\text{coord-cfg}}$ & Coordinate-guided velocity in the hierarchical fusion \\
$\mathbf{v}_t^{\text{final}}$ & Final fused velocity after text, coordinate, and image guidance \\
$s_{\text{text}}, s_{\text{img}}, s_{\text{coord}}$ & Guidance scales for text, image, and coordinate branches \\
$\hat{\mathbf{v}}_t$ & Renormalized guided velocity \\
$\mathcal{N}$ & Normalization domain (global or per-channel) \\
\bottomrule
\end{tabularx}
\end{table}

To improve layout controllability, we introduce a Coordinate-CFG mechanism that augments the text–image classifier-free guidance (CFG) used in Bagel~\cite{deng2025emerging} with an additional coordinate-guidance branch. When coordinate tokens are absent, the formulation reduces exactly to Bagel's original two-branch CFG. Quantitative results on COCO-Position are reported in Table~\ref{tab:coord_cfg_suppl}.

\begin{table*}[t]
\centering
\footnotesize
\setlength{\tabcolsep}{1.6pt}
\renewcommand{\arraystretch}{1.0}
\caption{Quantitative results of the effect of coordinate CFG on the COCO-Position benchmark.}
\label{tab:coord_cfg_suppl}
\begin{tabularx}{0.98\textwidth}{c|c|*{6}{>{\centering\arraybackslash}m{0.8cm}}|*{6}{>{\centering\arraybackslash}m{0.8cm}}|*{4}{>{\centering\arraybackslash}m{0.8cm}}}
\toprule
\multicolumn{2}{c|}{\textbf{Methods}} &
\multicolumn{6}{c|}{\textbf{Instance SR} $(\%) \uparrow$} &
\multicolumn{6}{c|}{\textbf{Image SR} $(\%) \uparrow$} &
\multicolumn{4}{c}{\textbf{Pos. Acc.} $(\%) \uparrow$} \\
\cmidrule(lr){1-2} \cmidrule(lr){3-8} \cmidrule(lr){9-14} \cmidrule(lr){15-18}
\textbf{method} & \textbf{coord cfg} &
$L_2$ & $L_3$ & $L_4$ & $L_5$ & $L_6$ & \textbf{Avg} &
$L_2$ & $L_3$ & $L_4$ & $L_5$ & $L_6$ & \textbf{Avg} &
mIoU & AP & AP50 & AP75 \\
\midrule
\multicolumn{2}{c|}{Bagel} &
19.1 & 15.2 & 13.8 & 11.4 & 11.0 & 13.1 &
3.7 & 0.6 & 0.0 & 0.0 & 0.0 & 0.9 &
23.1 & 0.7 & 3.2 & 0.2 \\
\midrule
\multirow{17}{*}{SFT}
& 0.2 & 68.4 & 68.5 & 62.7 & 58.3 & 56.1 & 61.1
      & 46.3 & 38.1 & 21.9 & 11.9 & 9.4  & 25.5
      & 56.0 & 21.4 & 45.0 & 18.4 \\
& 0.4 & 91.9 & 89.0 & 87.5 & 84.6 & 86.7 & 87.2
      & 83.8 & 70.6 & 60.0 & 50.0 & 46.9 & 62.3
      & 78.4 & 57.2 & 82.4 & 61.3 \\
& 0.6 & 94.7 & 92.9 & 90.3 & 87.0 & 89.8 & 90.2
      & 89.4 & 81.2 & 66.9 & 53.7 & 58.8 & 70.0
      & 82.3 & 64.8 & 86.7 & 70.2 \\
& 0.8 & 94.7 & 94.4 & 91.6 & 88.4 & 90.4 & 91.2
      & 89.4 & 83.8 & 70.6 & 58.8 & 62.5 & 73.0
      & 83.6 & 67.5 & 87.9 & 72.8 \\
& 1.0 & 95.3 & 93.8 & 91.7 & 90.2 & 91.6 & 92.0
      & 90.6 & 83.1 & 70.6 & 60.6 & 63.7 & 73.8
      & 84.4 & 69.2 & 88.1 & 74.4 \\
& 1.2 & 95.3 & 94.2 & 92.7 & 90.2 & 91.5 & 92.2
      & 90.6 & 83.1 & 71.9 & 60.6 & 64.4 & 74.1
      & 84.8 & 69.9 & 89.0 & 75.4 \\
& 1.4 & 95.0 & 94.2 & 93.0 & 90.6 & 92.4 & 92.6
      & 90.6 & 83.8 & 73.1 & 62.5 & 68.8 & 75.7
      & 85.2 & 70.5 & 89.1 & 75.5 \\
& 1.6 & 95.6 & 94.2 & 92.7 & 90.6 & 92.4 & 92.6
      & 91.9 & 83.1 & 73.1 & 63.7 & 68.8 & 76.1
      & 85.3 & 70.9 & 89.1 & 76.9 \\
& 1.8 & 96.3 & 94.4 & 92.5 & 91.2 & 92.8 & 92.9
      & 93.1 & 83.8 & 72.5 & 65.0 & 70.0 & 76.9
      & 85.6 & 71.7 & 89.4 & 77.4 \\
& 2.0 & 96.6 & 94.6 & 93.4 & 90.7 & 92.1 & 92.8
      & 93.1 & 85.0 & 75.6 & 64.4 & 66.9 & 77.0
      & 85.6 & 71.2 & 88.8 & 78.1 \\
& 2.2 & 95.9 & 95.2 & 93.3 & 90.9 & 92.5 & 93.0
      & 92.5 & 86.3 & 75.6 & 62.5 & 68.8 & 77.1
      & 85.5 & 71.2 & 89.2 & 76.8 \\
& 2.4 & 96.3 & 94.6 & 93.1 & 89.7 & 92.3 & 92.6
      & 92.5 & 85.0 & 75.0 & 61.9 & 65.0 & 75.9
      & 85.4 & 71.2 & 88.9 & 76.8 \\
& 2.6 & 96.3 & 94.6 & 92.5 & 89.7 & 91.9 & 92.3
      & 92.5 & 85.0 & 73.8 & 60.6 & 65.6 & 75.5
      & 85.2 & 70.9 & 89.1 & 76.7 \\
& 2.8 & 96.6 & 95.2 & 93.1 & 89.2 & 92.1 & 92.5
      & 93.1 & 86.9 & 75.6 & 60.6 & 66.9 & 76.6
      & 85.2 & 70.9 & 89.1 & 77.1 \\
& 3.0 & 96.9 & 94.8 & 93.4 & 90.6 & 91.6 & 92.7
      & 93.8 & 85.6 & 76.9 & 63.1 & 65.0 & 76.9
      & 85.4 & 71.1 & 89.3 & 77.1 \\
\bottomrule
\end{tabularx}
\end{table*}

\subsection{Notation}
For a noisy latent $\mathbf{x}_t$ at diffusion step $t$, we denote the velocity predictions of our model as
\[
\begin{aligned}
\mathbf{v}_t &= v_\theta(\mathbf{x}_t \mid \text{text}, \text{img}, \text{coord}), \\
\mathbf{v}_t^{\text{text-drop}} &= v_\theta(\mathbf{x}_t \mid \emptyset_{\text{text}}, \text{img}, \emptyset_{\text{coord}}), \\
\mathbf{v}_t^{\text{img-drop}} &= v_\theta(\mathbf{x}_t \mid \text{text}, \emptyset_{\text{img}}, \text{coord}), \\
\mathbf{v}_t^{\text{coord-drop}} &= v_\theta(\mathbf{x}_t \mid \text{text}, \text{img}, \emptyset_{\text{coord}}),
\end{aligned}
\]
where $v_\theta$ is the velocity prediction network and
$\emptyset_{\text{text-drop}}, \emptyset_{\text{img-drop}}, \emptyset_{\text{coord-drop}}$ denote \emph{dropping} (masking out) the text, image, and coordinate conditions, respectively.
The superscripts on $\mathbf{v}_t^{\text{text}}, \mathbf{v}_t^{\text{img}}, \mathbf{v}_t^{\text{coord}}$ indicate which modality is \emph{excluded} from the conditioning for that prediction, while the remaining modalities stay active (see the symbol table in Sec.~\ref{tab:symbol_table}).

\subsection{Hierarchical fusion}

Coordinate-CFG applies classifier-free guidance in a hierarchical manner, extending Bagel’s text–image CFG with an additional coordinate-guidance step.  
The guided velocity is computed as follows:
\begin{align}
\mathbf{v}_t^{\text{text-cfg}}
&= \mathbf{v}_t^{\text{text-drop}}
  + s_{\text{text}}\big(\mathbf{v}_t - \mathbf{v}_t^{\text{text-drop}}\big), \\
\mathbf{v}_t^{\text{coord-cfg}}
&= \mathbf{v}_t^{\text{coord-drop}}
  + s_{\text{coord}}\big(\mathbf{v}_t^{\text{text-cfg}} - \mathbf{v}_t^{\text{coord-drop}}\big), \\
\mathbf{v}_t^{\text{final}}
&= \mathbf{v}_t^{\text{img-drop}}
  + s_{\text{img}}\big(\mathbf{v}_t^{\text{coord-cfg}} - \mathbf{v}_t^{\text{img-drop}}\big).
\end{align}

This formulation is backward-compatible with the CFG used in Bagel \cite{deng2025emerging}. When the coordinate-guidance branch is disabled at sampling time (i.e., $\mathbf{v}_t^{\text{final}}$ is computed from the text and image branches without the intermediate coordinate-fusion step), the update rule reduces to the original two-branch text--image CFG parameterized by $s_{\text{text}}$ and $s_{\text{img}}$. Disabling the image-guidance branch as well further recovers the standard text-only CFG.

\subsection{Recommended Coordinate-CFG range}

We provide empirically validated coordinate-guidance scales based on MS-Bench \cite{wang2025msdiffusion} and COCO-Position evaluations.

\noindent\textbf{Text-to-Image.}
A coordinate guidance scale of $s_{\text{coord}} \in [0.6, 3.0]$ offers a favorable balance between enforcing layout constraints and preserving semantic flexibility. Lower values already ensure coarse spatial alignment, while higher values tighten the layout adherence at a moderate cost to sample diversity.

\noindent\textbf{Multi-Reference.}
For layout-controlled multi-reference generation, we recommend $s_{\text{coord}} \in [0.4, 1.6]$. Within this interval, increasing $s_{\text{coord}}$ improves positional stability and identity placement across multiple instances, whereas overly large values tend to over-constrain the layout and degrade visual fidelity.

\section{Dataset Construction}
\label{sec:dataset_construction_supp}

\begin{figure*}[t]
    \centering
    \includegraphics[width=1\linewidth]{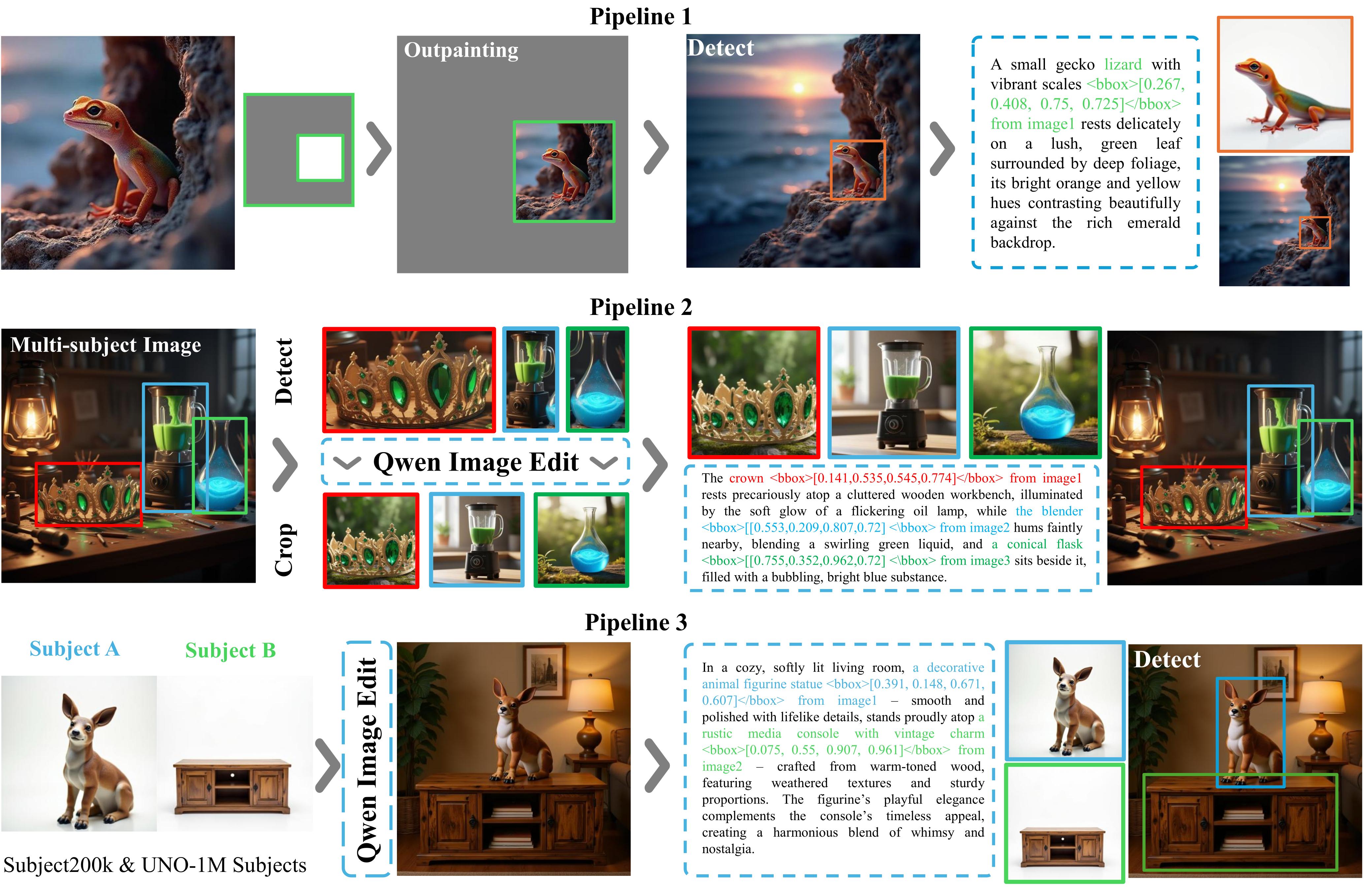}
    \caption{Overview of the reference-conditioned branch in ConsistCompose3M construction, comprising three pipelines. All annotations are represented using a unified inline \texttt{<bbox>} format.}
    \label{fig:data_pipeline}
\end{figure*}

We introduce ConsistCompose3M, a large-scale dataset for layout-controllable multi-instance image generation. Compared with existing resources, ConsistCompose3M provides substantially improved \emph{scale}, \emph{quality}, and \emph{adaptability}: it contains millions of diverse multi-instance scenes, includes identity-preserving samples filtered by similarity, and offers structured spatial and semantic supervision suitable for unified multimodal training. The corpus is organized into two complementary partitions—a \emph{layout-grounded text-to-image} split and a \emph{reference-conditioned} split designed for subject-preserving, layout-guided generation. Representative examples of the multi-subject data construction pipeline are shown in Figure~\ref{fig:data_pipeline}.

\medskip
\noindent\textbf{Unified representation.}
Each instance is annotated with an axis-aligned bounding box and a subject phrase. All spatial annotations are serialized inline using the format \texttt{<bbox>[$x_1,y_1,x_2,y_2$]</bbox>}, where $(x_1,y_1)$ and $(x_2,y_2)$ denote the top-left and bottom-right corners of the bounding box. Coordinates are normalized to $[0,1]$ using the image width $W$ and height $H$, and are rounded to three decimals. Throughout, we denote an image by $I$ and a normalized bounding box by $b = (x_1, y_1, x_2, y_2) \in [0,1]^4$.

\medskip
\noindent\textbf{Primary sources.}
ConsistCompose3M is constructed from four complementary sources.

\emph{(1) LayoutSAM.}
LayoutSAM~\cite{zhang2025creatilayout} provides multi-object scenes with instance-level bounding boxes and textual descriptions. We convert its annotations into our unified training format and, for the layout-grounded text-to-image split, additionally supply a global scene description for each image that summarizes all entities and their relations.

\emph{(2) UNO-1M.}
UNO-1M~\cite{wu2025less} is built on top of Objects365~\cite{shao2019objects365} and derives subject-centric data from its category set. We follow the class taxonomy used in UNO-1M \cite{wu2025less}and directly reuse its processed outputs—subject reference images and attribute-aware phrases—as paired sets $\{R_i\}$ and $\{s_i\}$. No additional subject categories are introduced.

\emph{(3) Subjects200K.}
Subjects200K~\cite{tan2025ominicontrol} is incorporated in the same manner, expanding both the coverage and appearance diversity of the subject bank.

\emph{(4) Virtual try-on data.}
To further enrich the reference-conditioned branch, we incorporate human–garment virtual try-on datasets such as VITON-HD ~\cite{choi2021viton} and DressCode-MR~\cite{chong2025fastfit}. Each sample consists of a reference garment image and a corresponding try-on image in which the garment is embedded within a layered human-body configuration, naturally inducing nested and occluded spatial relationships. We treat the garment image as the reference, annotate an axis-aligned bounding box for the garment region in the try-on image, and describe the scene using a standardized English virtual-try-on template.these samples provide complementary reference-conditioned supervision in which compositional constraints arise from body–garment interactions rather than explicit multi-object layouts.

\medskip
\subsection{Two data construction branches}
We construct two complementary data branches that share the same inline \texttt{<bbox>} representation but differ in whether reference images are provided.

\subsubsection{Layout-grounded T2I branch.}
\label{sec:dataset_construction_t2i_branch}
This branch is built from two sources:
(i) LayoutSAM-style corpora~\cite{zhang2025creatilayout} with given images, a \emph{global dense caption}, and instance-level spatial annotations, and
(ii) canonicalized samples derived from our reference-conditioned pipelines.
For (i), LayoutSAM \cite{zhang2025creatilayout} provides, for each image, a global dense caption $p^{\text{raw}}_{\text{dense}}$ and a set of entities with class labels and bounding boxes, where each instance is already mentioned in $p^{\text{raw}}_{\text{dense}}$.

We convert the global dense caption $p^{\text{raw}}_{\text{dense}}$ to a
layout-grounded LELG prompt $p_{\text{LELG}}$, preserving its narrative content
while explicitly encoding all annotated instances and their spatial coordinates,
using GPT-4o in an in-context learning configuration with the prompt design
shown in Fig.~\ref{fig:layout_t2i_prompt_b}. For each instance, we append its normalized bounding-box tag \texttt{<bbox>[$x_1,y_1,x_2,y_2$]</bbox>} immediately after the first occurrence of the corresponding phrase in $p_{\text{LELG}}$, producing layout-grounded T2I training pairs while transforming dense spatial annotations into inline, language-embedded supervision.

For (ii), we start from the outputs of Pipelines~1–3 (described below) and apply the deterministic conversion in Sec.~\ref{sec:dataset_construction_reference_branch}: we drop all reference-image inputs and remove any ``from image $k$'' indices from the tags, while keeping the scene sentence otherwise unchanged.

\subsubsection{Reference-conditioned branch}
\label{sec:dataset_construction_reference_branch}

This branch is built from UNO-1M~\cite{wu2025less} and Subjects200K~\cite{tan2025ominicontrol} subject banks and phrases processed by the three pipelines below (see Fig.~\ref{fig:data_pipeline}). Each sample consists of a scene sentence with inline \texttt{<bbox>} tags and a set of subject reference images $\{R_i\}$. The pipelines are explicitly designed for \emph{subject-consistent, layout-guided generation}. Through the canonicalization in Sec.~\ref{sec:dataset_construction_t2i_branch}, the same samples also contribute to the layout-grounded T2I branch. In addition, we use virtual try-on data from VITON-HD~\cite{choi2021viton} and DressCode-MR\cite{chong2025fastfit} as auxiliary sources, converting them into the same reference-conditioned format to further enrich the subject and appearance diversity of ConsistCompose3M.

\noindent\textbf{Pipeline 1: Layout Diversification via Controlled Outpainting.}
As shown in Pipeline~1 of Fig.~\ref{fig:data_pipeline}, this pipeline broadens the spatial distribution of bounding-box positions and scales in the single-reference setting using image pairs from Subjects200K~\cite{tan2025ominicontrol}, obtained through OmniControl’s side-by-side paired generation. For each sample, we first generate a dense caption $p^{\text{raw}}_{\text{dense}}$ for the reference image using GPT-4o~\cite{hurst2024gpt}. The reference image $R_i$ is then resized by a sampled side-length factor \(r \sim \mathcal{U}(0.6, 0.8)\) and placed onto a blank canvas at a random valid location, thereby defining a normalized bounding box \(b \in [0,1]^4\).

FLUX.1~Fill is subsequently applied to outpaint the surrounding regions, producing an initial scene completion while preserving the pasted reference. A feathered binary mask is finally used to blend the pasted region with the outpainted context, yielding a coherent image \(I\). Because subjects in Subjects200K reference image are typically centered, the random rescaling and repositioning effectively counteract this center bias and support a wider range of spatial configurations. The resulting box \(b\) is recorded as the inline \texttt{<bbox>} annotation, providing diverse layouts and informative supervision for layout-aware generation.

\begin{figure}[t]
    \centering
    \includegraphics[width=\linewidth]{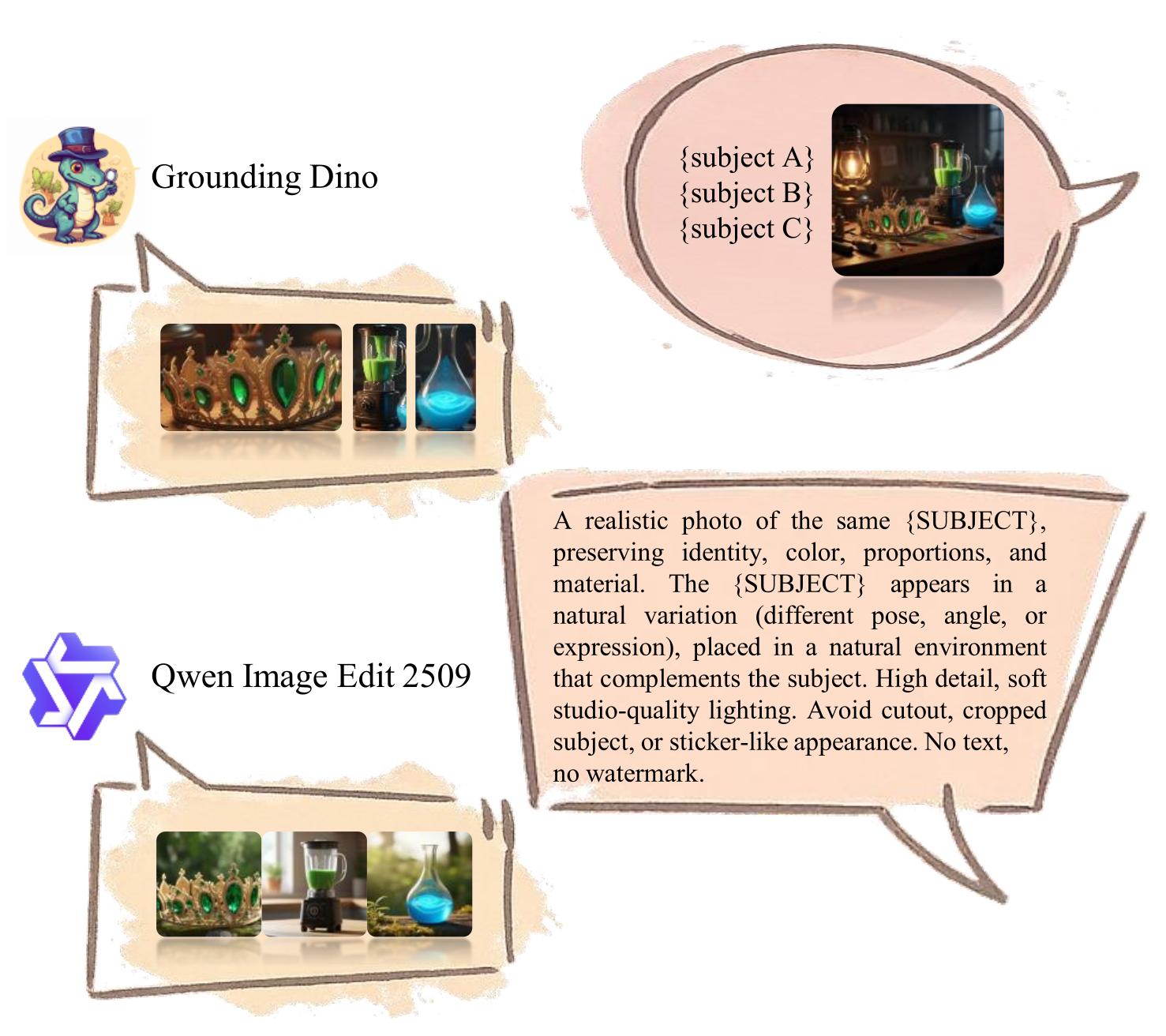}
    \caption{Crop-based single-reference refinement. Starting from a
    multi-subject scene, a subject crop is extracted using its detected bounding
    box and refined by Qwen Image Edit in a subject-to-image manner,
    producing an identity-preserving single-reference sample.}
    
    \label{fig:prompt_chat}
\end{figure}

\medskip
\noindent\textbf{Pipeline 2: Crop-based single-reference refinement.}
As illustrated in Pipeline~2 of Fig.~\ref{fig:data_pipeline}, this pipeline refines single-reference samples by extracting and enhancing subject crops from multi-subject scenes. We begin by applying GroundingDINO to detect subject-specific bounding boxes, retaining boxes with area ratios in $[0.20,0.60]$, removing high-IoU duplicates, and requiring at least three valid subjects per image. For each surviving subject–box pair $(R_i, b)$, we extract the corresponding crop $I|_b$ and retain candidates that satisfy the CLIP-T constraint $c(I|_b, R_i) \ge 0.25$.
The selected crop is then refined using Qwen Image Edit as shown in Fig.~\ref{fig:prompt_chat}, yielding a generated image $I'$. We subsequently filter the generation by comparing $I'$ directly with the original crop $I|_b$ using CLIP-I/CLIP-T scores, and convert the resulting valid candidate into an inline \texttt{<bbox>} tag via ICBP.

\medskip
\noindent\textbf{Pipeline 3: Multi-Reference Guided Scene Synthesis.}
As shown in Pipeline~3 of Fig.~\ref{fig:data_pipeline}, we construct multi-subject scenes by sampling $K\!\in\!\{2,\dots,4\}$ distinct subject--reference pairs $\{(s_i, R_i)\}$ from UNO-1M (Object365-derived) and Subjects200K. We then query GPT-4o with the instruction template in Fig.~\ref{fig:english_prompt_template} to obtain a concise global dense prompt $p^{\text{raw}}_{\text{dense}}$ that includes each subject phrase $s_i$ verbatim within the description.

Qwen Image Edit synthesizes a multi-subject image $I$ from the image references $\{R_i\}$ and the text prompt $p^{\text{raw}}_{\text{dense}}$.  Instead of directly pasting reference crops onto a canvas, the model re-renders the subjects coherently within the scene, substantially reducing trivial copy--paste artifacts commonly observed in multi-reference generation.

To link each subject reference to a concrete location in $I$ and filter out erroneous generations, we run GroundingDINO on $I$ for each subject phrase to obtain subject-specific candidate boxes $\{b_j\}$. For every subject $s_i$ with its reference image $R_i$ and every candidate box $b_j$, we crop the corresponding region $I|_{b_j}$ and compute three scores: (i) a CLIP text–image similarity $s^{\mathrm{T}}_{ij}$ between $I|_{b_j}$ and the subject phrase $s_i$, (ii) a CLIP image–image similarity $s^{\mathrm{I}}_{ij}$ between $I|_{b_j}$ and the reference image $R_i$, and (iii) a GroundingDINO detection score $s^{\mathrm{D}}_{ij}$ associated with $b_j$. After normalizing all scores to $[0,1]$, we form a combined similarity

\begin{equation}
S_{ij}
=
\alpha\, s^{\mathrm{T}}_{ij}
+
\beta\, s^{\mathrm{D}}_{ij}
+
\gamma\, s^{\mathrm{I}}_{ij},
\qquad
C_{ij} = 1 - S_{ij},
\label{eq:combined_cost}
\end{equation}
where $\alpha,\beta,\gamma \ge 0$ and $\alpha + \beta + \gamma = 1$, and $C_{ij}$ is the cost of assigning subject $i$ to box $j$.
We then solve the linear assignment problem
\begin{equation}
\begin{aligned}
\min_{\mathbf{X}} \quad &
\sum_{i=1}^{M} \sum_{j=1}^{N} C_{ij} X_{ij} \\
\text{s.t.}\quad &
\mathbf{X} \in \{0,1\}^{M\times N}, \\
& \sum_{j=1}^{N} X_{ij} = 1,\quad i=1,\dots,M, \\
& \sum_{i=1}^{M} X_{ij} \le 1,\quad j=1,\dots,N.
\end{aligned}
\label{eq:hungarian}
\end{equation}
via the Hungarian algorithm, where $M$ and $N$ denote the numbers of subjects and detected boxes, respectively, and $\mathbf{X}$ is the assignment matrix. This yields a unique bounding box in $I$ for each subject, or rejects the scene if a complete, non-duplicated matching is impossible. We further discard a scene if any matched pair fails per-pair quality checks (e.g., CLIP-T, CLIP-I, or detection scores falling below preset thresholds), so this matching step simultaneously aligns references to spatial positions in the synthesized image and aggressively filters out low-quality or mismatched generations.

Finally, the three pipelines collectively yield, for each constructed sample, a scene-level dense description and the corresponding subject–bounding-box associations. We then convert the raw dense sentence $p^{\text{raw}}_{\text{dense}}$ into a layout-grounded, multi-reference prompt $p_{\text{LELG}}$ by inserting explicit bounding-box and source-image tags. For each matched subject, we take its normalized bounding box $[x_1,y_1,x_2,y_2]$ together with the index of its reference image and employ GPT-4o~\cite{hurst2024gpt} with the instruction template in Fig.~\ref{fig:bbox_insertion_prompt} to insert a tag of the form \texttt{<bbox>[$x_1,y_1,x_2,y_2$]</bbox> from image $k$} immediately after the first noun-phrase mention of that subject in $p^{\text{raw}}_{\text{dense}}$. By integrating reference-guided synthesis, detection-and-matching–based filtering, and LLM-driven tag insertion, these pipelines produce layout-grounded samples exhibiting rich nested and attributive configurations.

\section{Supplementary Results and Detailed Analysis}

\begin{figure*}
  \centering
  \includegraphics[width=1.0\linewidth]{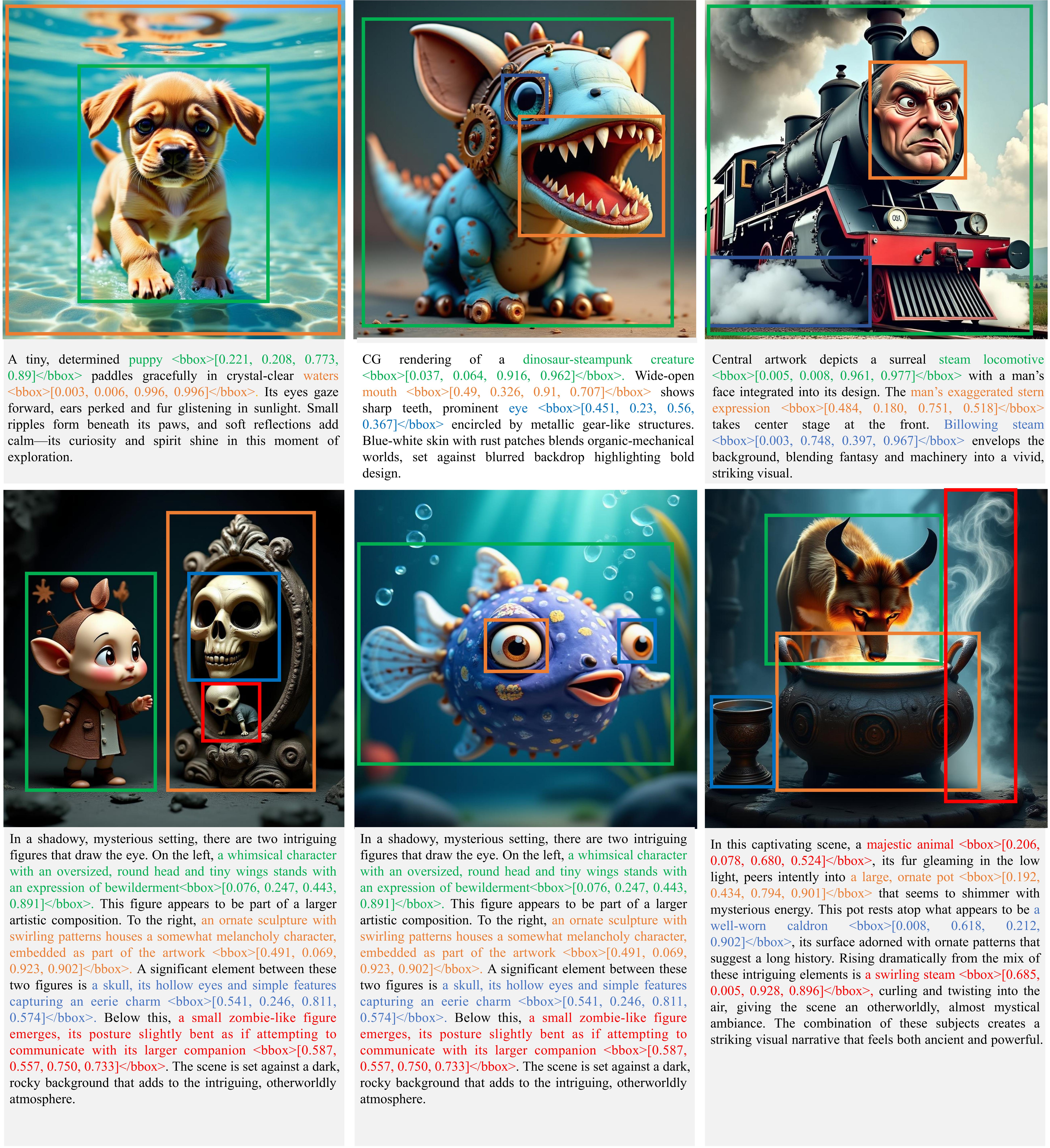}
  \vspace{-4.5mm}
  \caption{
  Qualitative examples of ConsistCompose for layout-controllable multi-instance generation under the ICBP paradigm.
  The model performs text-driven T2I generation with faithful prompt alignment and high visual fidelity, and synthesizes subjects with precise attributes and spatial arrangements.
  }
  \label{fig:gallery}
  \vspace{-2.5mm}
\end{figure*}

\begin{figure*}[t]
    \centering
    \includegraphics[width=\linewidth]{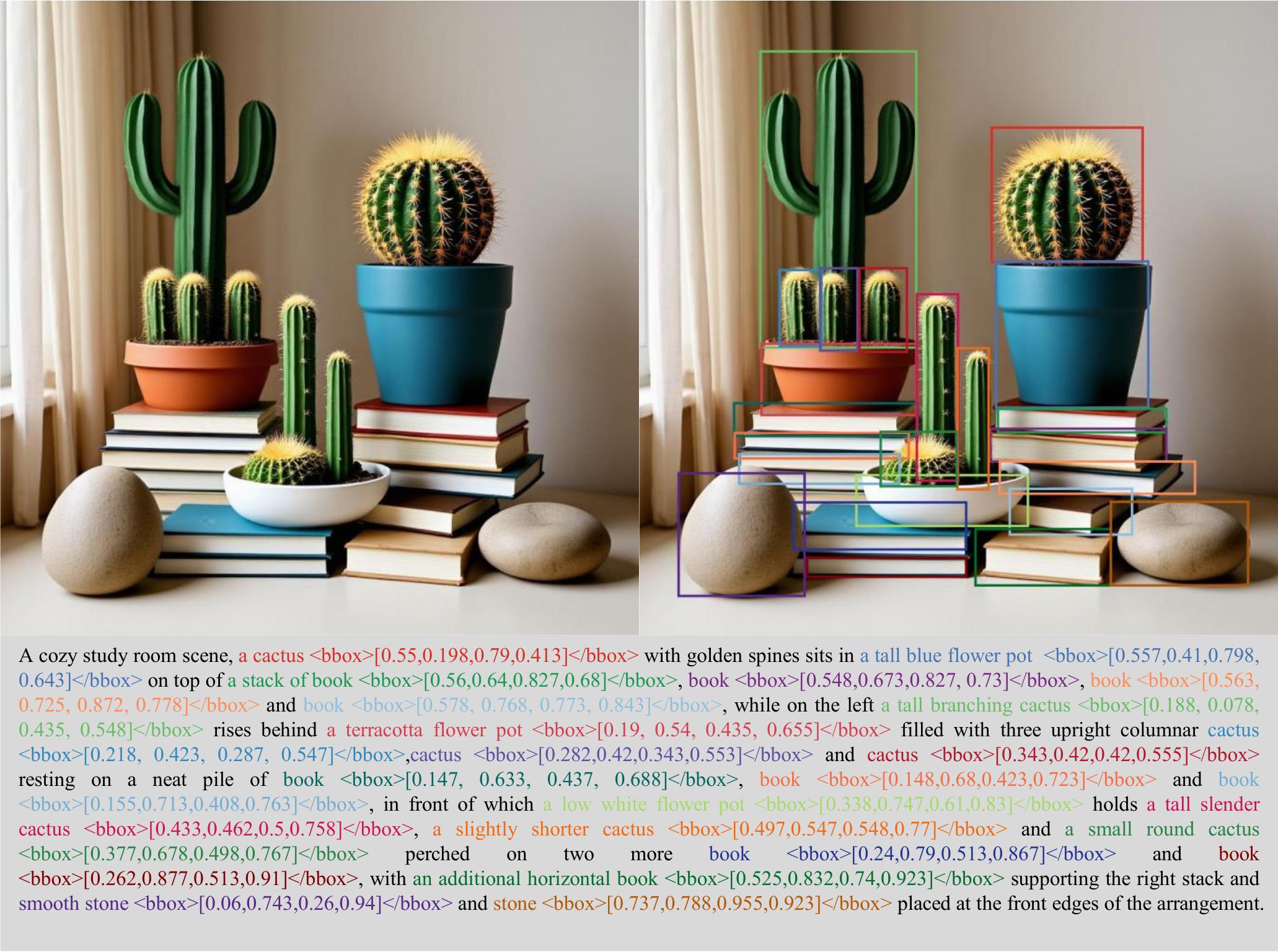
}
    \caption{Layout-grounded text-to-image generation in an extremely dense multi-instance scene with overlapping and nested subjects.}
    \label{fig:layout-grounded_t2i_complex_scene}
\end{figure*}

\noindent\textbf{Layout-Grounded Text-to-Image Generation.}
We provide additional qualitative results to complement the main paper.
\cref{fig:gallery} presents extended examples of layout-grounded multi-instance generation under the ICBP paradigm.
Moreover, \cref{fig:layout-grounded_t2i_complex_scene} showcases more challenging, heavily constrained scenarios with overlapping and closely interacting subjects, where our method continues to preserve object count, fine-grained attributes (e.g., color and pose), and precise spatial arrangements.
Across these cases, the model maintains high visual fidelity and a consistent global style, further substantiating the quantitative improvements.

\begin{table}[t]
  \centering
  \caption{Quantitative results on GenEval. * indicates methods using an LLM-based rewriter.}
  \setlength{\tabcolsep}{2.5pt}
  \renewcommand{\arraystretch}{1.1}
  \footnotesize

  \begin{tabularx}{\columnwidth}{
    @{}l
    *{6}{>{\hsize=0.95\hsize\centering\arraybackslash}X}
    >{\hsize=1.3\hsize\centering\arraybackslash}X@{}
  }
    \toprule
    \textbf{Model} &
    \textbf{Single} &
    \textbf{Two} &
    \textbf{Count} &
    \textbf{Color} &
    \textbf{Pos.} &
    \textbf{Attr.} &
    \textbf{Overall$\uparrow$}
    \\
    \midrule
    SDXL          & 0.98 & 0.74 & 0.39 & 0.85 & 0.15 & 0.23 & 0.55 \\
    FLUX.1-dev    & 0.99 & 0.82 & 0.78 & 0.73 & 0.19 & 0.46 & 0.66 \\
    Show-o        & 0.98 & 0.80 & 0.67 & 0.83 & 0.32 & 0.50 & 0.68 \\
    Janus         & 0.97 & 0.68 & 0.31 & 0.84 & 0.46 & 0.42 & 0.61 \\ 
    Janus Pro     & 0.99 & 0.89 & 0.59 & 0.90 & 0.79 & 0.66 & 0.80 \\
    Bagel$^*$     & 0.98 & 0.95 & 0.84 & 0.93 & 0.72 & 0.73 & 0.86 \\
    \rowcolor[HTML]{E6F2FF}
    Alignment$^*$ & 0.98 & 0.97 & 0.77 & 0.93 & 0.84 & 0.80 & 0.88 \\
    \rowcolor[HTML]{E6F2FF}
    SFT$^*$   & 0.98 & 0.95 & 0.79 & 0.94 & 0.84 & 0.79 & 0.88 \\
    \bottomrule
  \end{tabularx}
  \label{tab:genevel_quantitative}
\end{table}

\begin{table}[t]
  \centering
  \caption{Editing performance on GEdit-Bench-EN.}
  \begin{tabular}{lccc}
    \toprule
    \textbf{Model} & \textbf{SC$\uparrow$} & \textbf{PQ$\uparrow$} & \textbf{OS$\uparrow$} \\
    \midrule
    Bagel Base             & 6.94 & 6.73 & 6.36 \\
    \rowcolor[HTML]{E6F2FF}
    Ours (w/o Coord Data)  & 6.09 & 6.72 & 5.87 \\
    \rowcolor[HTML]{E6F2FF}
    Ours (w/ Coord Data)   & 6.32 & 6.82 & 5.78 \\
    \bottomrule
  \end{tabular}
  \label{tab:gedit}
\end{table}

\noindent\textbf{General Capabilities.}
We study whether introducing ICBP coordinate data for layout grounding affects the model’s general capabilities in generation, image editing, and multimodal understanding.

For generation, we evaluate fine-grained control on GenEval (Table~\ref{tab:genevel_quantitative}). Our Alignment$^*$ and SFT$^*$ (with an LLM-based rewriter) achieve an overall score of 0.88, with strong position control (0.84) and attribute consistency (0.79--0.80), indicating that precise layout adherence does not compromise control granularity.

For editing, to control for data/recipe confounders, we evaluate on GEdit-Bench-EN (Table~\ref{tab:gedit}) with three settings: (i) \emph{Bagel Base}, the original pretrained model; (ii) \emph{Ours (w/o Coord Data)}, trained with the same data mixture and optimization recipe but without injecting ICBP coordinate data; and (iii) \emph{Ours (w/ Coord Data)}, trained identically while including such coordinate data. Both trained variants show a mild drop compared to Bagel Base, while \emph{w/ Coord Data} is comparable to \emph{w/o Coord Data} under matched data and optimization, suggesting that ICBP coordinate data does not introduce additional regression on general editing.

For understanding, we report results on general multimodal benchmarks (Table~\ref{tab:mme_llm_comparison}). Alignment variants (with/without coord data) achieve $\sim$81.4 on MMBench, comparable to InternVL3-8B (81.7)~\cite{Zhu2025InternVL3EA} and Bagel (81.4), validating no degradation in basic visual understanding. Notably, on MMMU, introducing coord data improves Alignment from 39.4 (w/o coord data) to 42.3 (w/ coord data), suggesting coord data may benefit spatial reasoning.

\begin{table}[t]
  \centering
  \caption{Quantitative results on general understanding benchmarks.}
  \label{tab:mme_llm_comparison}
  \setlength{\tabcolsep}{4pt}
  {\small
  \begin{tabular}{@{}lcc@{}}
    \toprule
    \textbf{Model} & \textbf{MMBench$\uparrow$} & \textbf{MMMU$\uparrow$} \\
    \midrule
    Qwen2.5-VL-3B-Instruct     & 77.4 & 53.1 \\
    Qwen2.5-VL-7B-Instruct     & 82.6 & 58.6 \\
    InternVL3-8B               & 81.7 & 65.6 \\
    Bagel                      & 81.4 & 46.4 \\
    \rowcolor[HTML]{E6F2FF}
    Alignment (w/o Coord Data) & 81.5 & 39.4 \\
    \rowcolor[HTML]{E6F2FF}
    Alignment (w/ Coord Data)  & 81.4 & 42.3 \\
    \rowcolor[HTML]{E6F2FF}
    SFT (w/ Coord Data)        & 75.3 & 40.2 \\
    \bottomrule
  \end{tabular}
  }
\end{table}

\begin{table}[t]
  \centering
  \caption{Quantitative results on DreamBench (Single/Multi)}
  \resizebox{\linewidth}{!}{
    \begin{tabular}{lcccccc}
      \toprule
      \multirow{2}{*}{\makecell[c]{\textbf{Method}}} & \multicolumn{3}{c}{\textbf{DreamBench Single}} & \multicolumn{3}{c}{\textbf{DreamBench Multi}} \\
      \cmidrule(lr){2-4} \cmidrule(lr){5-7}
      & \textbf{DINO} & \textbf{CLIP-I} & \textbf{CLIP-T} & \textbf{DINO} & \textbf{CLIP-I} & \textbf{CLIP-T} \\
      \midrule
      Step1X-Editing & 0.616 & 0.779 & 0.314 & -- & -- & -- \\
      OmniGen                           & 0.554 & 0.746 & 0.322 & 0.441 & 0.692 & 0.341 \\
      OmniGen2                          & 0.671 & 0.791 & 0.312 & 0.459 & 0.698 & 0.333 \\
      UNO                               & 0.661 & 0.796 & 0.304 & 0.491 & 0.715 & 0.323 \\
      Flux Kontext dev                  & 0.687 & 0.806 & 0.310 & 0.500 & 0.712 & 0.328 \\
      Qwen Image Edit                   & 0.638 & 0.780 & 0.326 & 0.458 & 0.697 & 0.342 \\
      GPT4O                             & 0.687 & 0.801 & 0.310 & 0.529 & 0.725 & 0.324 \\
      \rowcolor[HTML]{E6F2FF}
      Alignment (w/o Coord Data)        & 0.565 & 0.757 & 0.325 & 0.456 & 0.699 & 0.342 \\
      \rowcolor[HTML]{E6F2FF}
      Alignment (w/  Coord Data)        & 0.611 & 0.770 & 0.320 & 0.473 & 0.701 & 0.337 \\
      \rowcolor[HTML]{E6F2FF}
      SFT  (w/ Coord Data)              & 0.677 & 0.792 & 0.314 & 0.506 & 0.703 & 0.335 \\
      \bottomrule
    \end{tabular}
  }
  \label{tab:dreambench_perf}
\end{table}

\begin{figure*}
  \centering
  \includegraphics[width=1.0\linewidth]{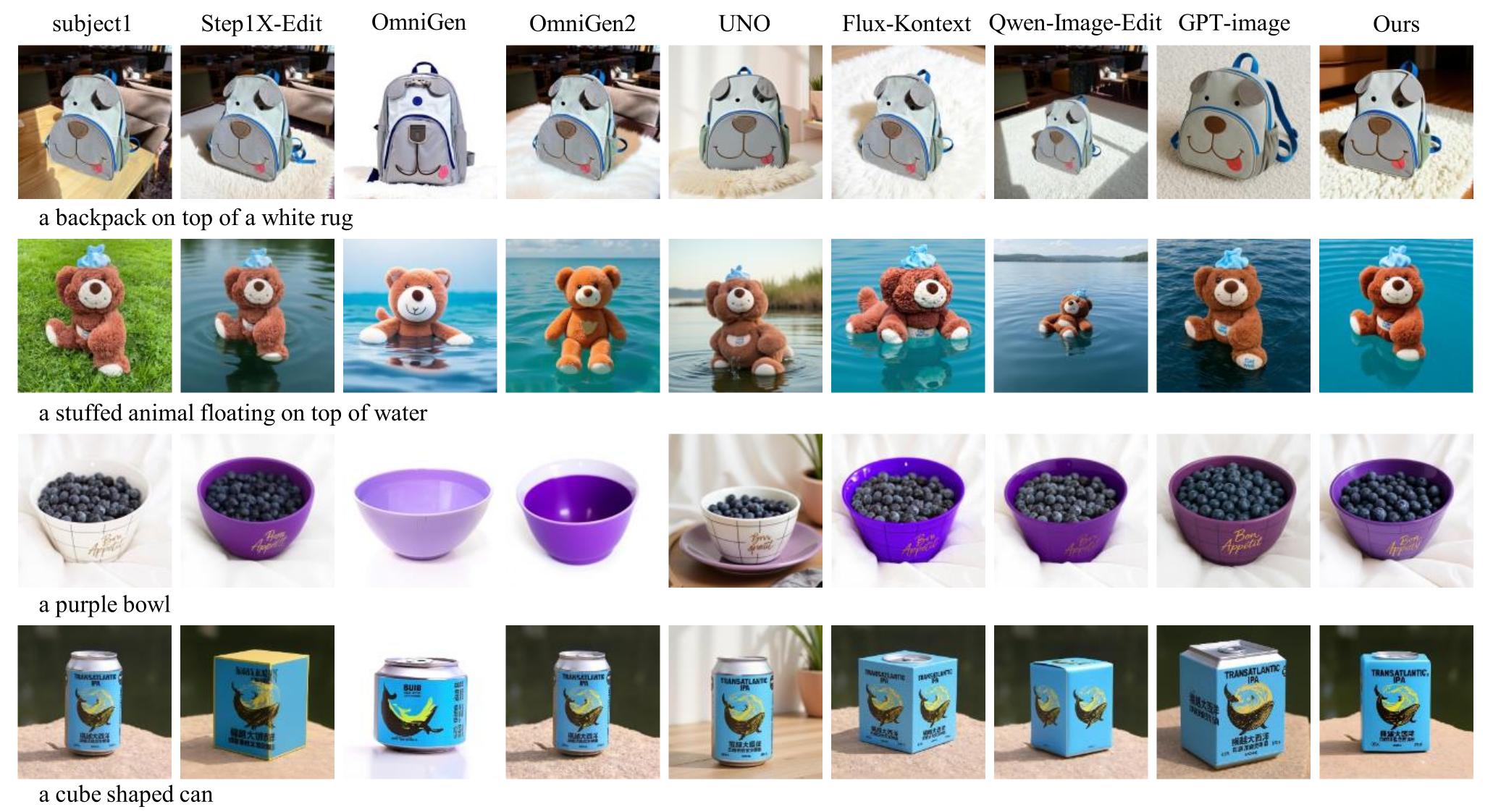}
  \vspace{-4.5mm}
  \caption{
  Qualitative comparison on DreamBench single-subject generation.
  Across diverse prompts, ConsistCompose preserves subject identity and follows the textual description comparably to state-of-the-art methods.
  }
  \label{fig:db_single}
  \vspace{-2.5mm}
\end{figure*}

\begin{figure*}
  \centering
  \includegraphics[width=1.0\linewidth]{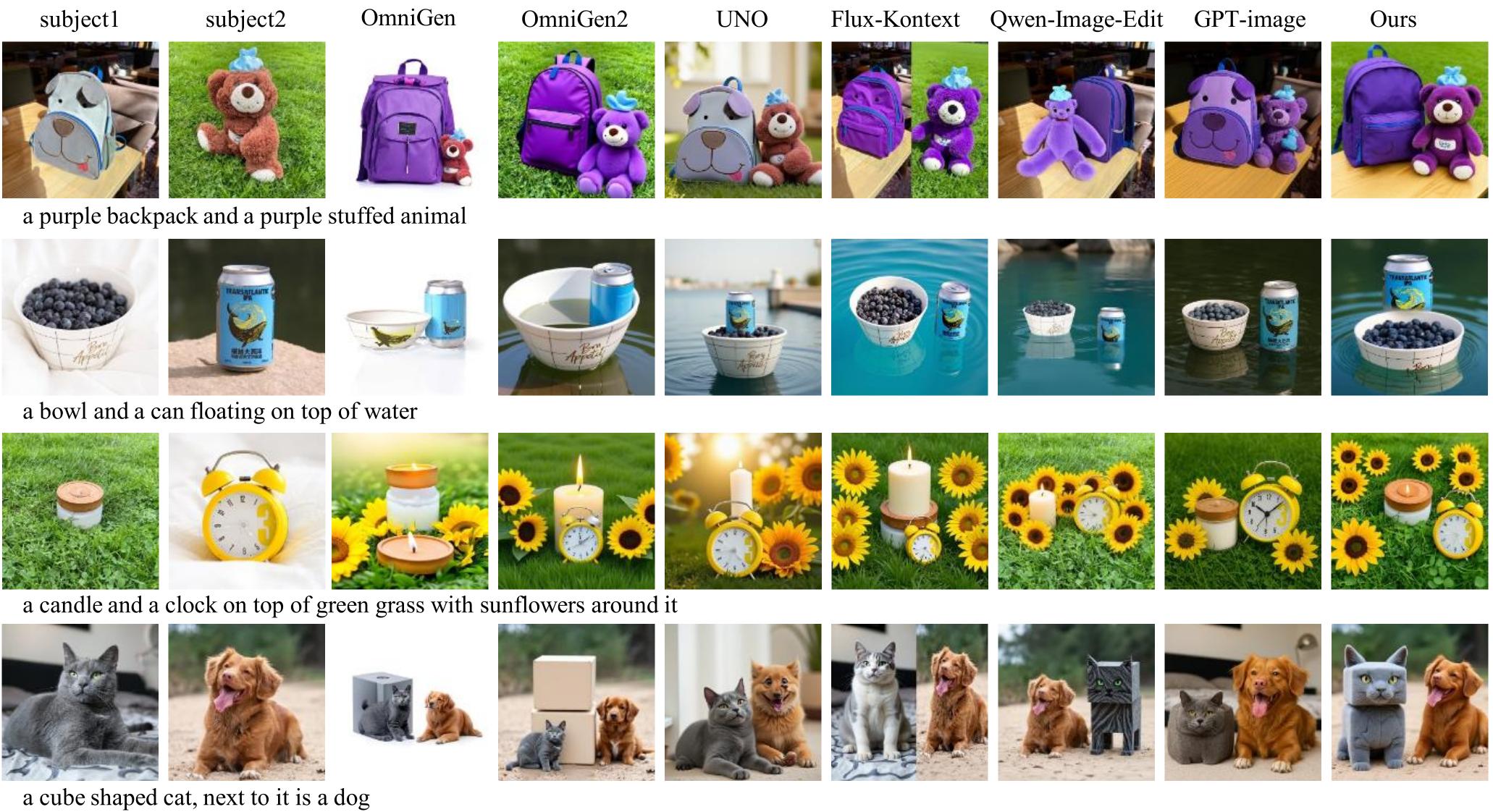}
  \vspace{-4.5mm}
  \caption{
  Qualitative comparison on DreamBench multi-subject generation.
  In more challenging multi-subject and multi-scenario settings, ConsistCompose maintains consistent subject appearance and prompt faithfulness on par with state-of-the-art methods.
  }
  \label{fig:db_multi}
  \vspace{-2.5mm}
\end{figure*}

\noindent\textbf{Extended DreamBench Results.}
To assess generalization beyond layout control, we additionally report DreamBench results (\cref{tab:dreambench_perf}, \cref{fig:db_single,fig:db_multi}). \emph{Alignment (w/ Coord Data)} outperforms its non-coordinate counterpart (0.611 vs.\ 0.565 in the DreamBench single-subject DINO identity metric), and \emph{SFT (w/ Coord Data)} further reaches 0.677 for single-subject and 0.506 for multi-subject generation, surpassing baselines such as OmniGen2~\cite{wu2025omnigen2explorationadvancedmultimodal} and narrowing the gap to top-performing models (Flux Kontext dev~\cite{labs2025flux}, GPT-4o~\cite{hurst2024gpt}). Qualitative examples in \cref{fig:db_single,fig:db_multi} corroborate these gains, showing strong identity preservation and prompt adherence for single subjects, as well as consistent identities with natural interactions in multi-subject scenes. Together, these results indicate that our method attains strong layout control without compromising general understanding or broad generative capability.

\textbf{\begin{table}[t]
\centering
\scriptsize
\setlength{\tabcolsep}{4pt}
\caption{COCO-Position: comparison with general image generation models. Instance/Image success ratio (Avg, \%) and position accuracy (\%).}
\begin{tabular}{lcccccc}
\toprule
\textbf{Method} & \textbf{Inst. SR} & \textbf{Img. SR} & \textbf{mIoU} & \textbf{AP} & \textbf{AP50} & \textbf{AP75} \\
\midrule
QwenImage & 7.9 & 0.3 & 18.6 & 0.3 & 1.5 & 0.1 \\
Nano Banana & 11.2 & 0.1 & 22.8 & 0.5 & 2.4 & 0.1 \\
\rowcolor[HTML]{E6F2FF}
Ours & 92.6 & 76.1 & 85.3 & 70.9 & 89.1 & 76.9 \\
\bottomrule
\end{tabular}
\label{tab:coco_position_compact_supp_gen}
\end{table}}

\noindent\textbf{Comparison with general image generation models.} We compare our model with general-purpose image generation models on COCO-Position (Table~\ref{tab:coco_position_compact_supp_gen}).
Despite strong text-to-image quality, these general models struggle to satisfy explicit multi-instance spatial constraints, resulting in very low instance/image success rates and poor localization accuracy.
In contrast, our unified model achieves substantially higher success rates and more accurate localization across all metrics.

\begin{figure}[h]
    \centering
    \includegraphics[width=\linewidth]{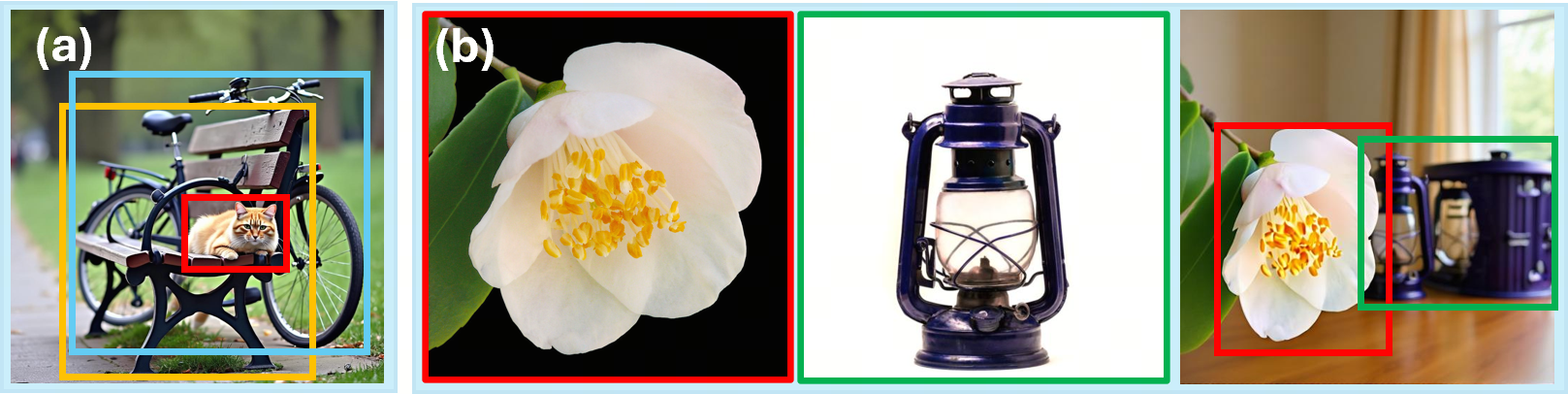}
    \caption{Interaction inconsistency \& unintended additions.}
    \label{fig:action_table}
\end{figure}

\noindent\textbf{Failure Case Analysis.}
As illustrated in Figure~\ref{fig:action_table}, failures arise when the model cannot infer a geometry-consistent global pose and occlusion/contact topology from the layout.
This leads to two typical error patterns:
(a) \emph{inconsistent interactions} between objects, and
(b) \emph{unintended additions} that only satisfy local spatial constraints while violating global consistency.

\begin{figure*}[b]
\centering
\footnotesize
\begin{tcolorbox}[
    width=\linewidth,
    colback=gray!8,
    colframe=gray!40,
    fontupper=\color{black},
    arc=2mm,
    boxrule=0.3pt,
    left=4mm,right=4mm,top=3mm,bottom=3mm
]
\textbf{Requirements}\\[0.2em]
\textbf{1.~When a subject appears multiple times in the original prompt:}\\
\quad (a) If referring to the same instance, merge mentions and place tags after the first occurrence.\\
\quad (b) If referring to different instance, tag each occurrence separately.\\
\quad (c) Add count prefixes (e.g., ``\texttt{2 backpacks}'') when appropriate for clarity.\\[0.4em]
\textbf{2.~For subjects with single bbox:} place the tag immediately after the first occurrence of the subject, e.g., \texttt{[subject] <bbox>[...]</bbox>}.\\[0.4em]
\textbf{3.~For subjects with multiple bboxes:} ensure the count matches the number of bboxes provided,   e.g., \texttt{[count] [subject] <bbox>[...]</bbox>, \dots, <bbox>[...]</bbox>}.\\[0.4em]
\textbf{4.~Make minimal necessary modifications only to}:\\
\quad (a) Improve clarity when adding multiple bbox tags.\\
\quad (b) Resolve redundancy in repeated subject mentions.\\
\quad (c) Adjust grammar for coherence after tag insertion.\\[0.4em]
\textbf{5.~Preserve all original information. including:}\\
\quad (a) Specific details and attributes.\\
\quad (b) Overall meaning and context.\\
\quad (c) Sentence structure when possible.\\[0.4em]
\textbf{6. Output Format (Critical):}~Only return the modified prompt with bbox tags. Do \emph{not} add any extra content, such as titles (e.g., ``Modified Prompt''), comments, or explanations.\\[0.8em]
\textbf{Example 1}\\
\textbf{Original:} ``A hiker wears a backpack. The backpack is black. He also carries a water bottle.''\\
\textbf{Tags:} ``backpack'' (Details: a large black hiking backpack) $\rightarrow$ \texttt{1 backpack <bbox>[0.3,0.4,0.5,0.7]</bbox>}\\
\textbf{Expected Output:} ``A hiker wears a backpack \textless bbox\textgreater[0.3,0.4,0.5,0.7]\textless/bbox\textgreater{} which is black. He also carries a water bottle.''\\[0.8em]
\textbf{Example 2}\\
\textbf{Original:} ``Dogs play in the park. The dogs chase each other happily.''\\
\textbf{Tags:} ``dogs'' $\rightarrow$ \texttt{3 dogs <bbox>[0.1,0.2,0.3,0.4]</bbox>, <bbox>[0.5,0.6,0.7,0.8]</bbox>, <bbox>[0.2,0.3,0.4,0.5]</bbox>}\\
\textbf{Expected Output:} ``3 dogs \textless bbox\textgreater[0.1, 0.2, 0.3, 0.4]\textless/bbox\textgreater, \textless bbox\textgreater[0.5, 0.6, 0.7,  0.8]\textless/bbox\textgreater, \textless bbox\textgreater[0.2, 0.3, 0.4, 0.5]\textless/bbox\textgreater{} play in the park. The dogs chase each other happily.''
\end{tcolorbox}
\caption{Requirements used in the prompt template for
layout-grounded T2I branch. This part specifies how to insert inline
\texttt{<bbox>} tags into a dense caption $p^{\text{raw}}_{\text{dense}}$.}

\label{fig:layout_t2i_prompt_a}
\end{figure*}

\begin{figure*}[t]
\centering
\begin{tcolorbox}[
    width=\linewidth,
    colback=gray!10,
    colframe=gray!40,
    fontupper=\footnotesize\color{black},
    arc=2mm,
    boxrule=0.3pt,
    left=4mm,right=4mm,top=3mm,bottom=3mm
]

\textbf{Global Dense Caption:}\\[0.2em]
This is a photo showcasing a corner of a city street, with a row of \textbf{orange-red buildings} in the background. In the foreground, there is a green trash can with a \textbf{vintage-style advertisement} on it, featuring a man and a woman in a classic movie poster style. Next to the trash can is a \textbf{wooden bench}, with a sign displaying a schedule of performances on a stand in front of it. The entire scene is illuminated by natural light, creating a tranquil and nostalgic atmosphere.\\[0.6em]
\textbf{Instance with Spatial Annotation:}\\[0.2em]
\texttt{'wooden bench'} $\rightarrow$ \texttt{<bbox>[0.497,0.725,0.998,0.995]</bbox>}\\
\texttt{'vintage-style advertisement'} $\rightarrow$ \texttt{<bbox>[0.2,0.326,0.386,0.813]</bbox>}\\
\texttt{'orange-red buildings'} $\rightarrow$ \texttt{<bbox>[0.002,0.003,0.997,0.876]</bbox>}\\[0.8em]
\textbf{Requirements:}\\
(omitted here for brevity; see main prompt in Fig.~\ref{fig:layout_t2i_prompt_a})\\[0.8em]
\textbf{Resulting LELG prompt:}\\[0.3em]
\begin{tcolorbox}[
    colback=white,
    colframe=gray!60,
    boxrule=0.3pt,
    left=2mm,right=2mm,top=2mm,bottom=2mm,
    arc=1mm,
    enhanced,
    fontupper=\scriptsize\color{black}
]
This is a photo showcasing a corner of a city street, with a row of orange-red buildings 
\textless bbox\textgreater[0.002,0.003,0.997,0.876]\textless/bbox\textgreater{} in the background. 
In the foreground, there is a green trash can with a vintage-style advertisement 
\textless bbox\textgreater[0.2,0.326,0.386,0.813]\textless/bbox\textgreater{} on it, featuring a man and a woman in a classic movie poster style. 
Next to the trash can is a wooden bench 
\textless bbox\textgreater[0.497,0.725,0.998,0.995]\textless/bbox\textgreater{}, with a sign displaying a schedule of performances on a stand in front of it. 
The entire scene is illuminated by natural light, creating a tranquil and nostalgic atmospher.
\end{tcolorbox}
\end{tcolorbox}
\caption{Prompt template for 
layout-grounded T2I branch. The figure shows the input dense caption
$p^{\text{raw}}_{\text{dense}}$, its instance--bbox annotations, and the
resulting layout-grounded sentence $p_{\text{LELG}}$ with inline
\texttt{<bbox>} tags.}

\label{fig:layout_t2i_prompt_b}
\end{figure*}

\begin{figure*}[t]
\centering
\begin{tcolorbox}[
    width=\textwidth,
    colback=gray!10,
    colframe=gray!40,
    arc=2mm,
    boxrule=0.3pt,
    left=4mm,right=4mm,top=3mm,bottom=3mm
]
\footnotesize
Create a detailed English scene description that includes the following exact subjects:\\
\{subject list\}\\[0.8em]

\textbf{IMPORTANT REQUIREMENTS:}
\begin{enumerate}[leftmargin=1.6em,itemsep=0.2em,topsep=0.2em]
  \item Use the exact subject names as provided --- \emph{do not} paraphrase or rename them.
  \item Show clear interactions between the subjects.
  \item Add specific environment details (lighting, setting, time of day).
  \item Use vivid, descriptive language suitable for image generation.
  \item Keep it concise (1--2 sentences); no explanations, just the scene.
\end{enumerate}

\vspace{0.6em}
\textbf{Example:}\\
\texttt{["cat", "yarn ball"]}\\[0.3em]
\textbf{Example output:}\\
``The cat plays with the yarn ball on a sunny windowsill, batting at the colorful strands with its paw.''

\end{tcolorbox}
\caption{Prompt template for generating $p^{\text{raw}}_{\text{dense}}$ (Reference-conditioned Branch)}
\label{fig:english_prompt_template}
\end{figure*}

\begin{figure*}[t]
\centering
\begin{tcolorbox}[
    width=\textwidth,
    colback=gray!10,
    colframe=gray!40,
    arc=2mm,
    boxrule=0.3pt,
    left=4mm,right=4mm,top=3mm,bottom=3mm
]
\footnotesize
Your task is to modify the original English prompt by inserting a bbox tag \textbf{after} each specified subject, \textbf{without} changing any other content of the original prompt.\\[0.8em]

\textbf{Original:}\\
\{original\_prompt\}\\[0.6em]

\textbf{Tags (keep original names, insert tag immediately after the subject):}\\
\{Subjects to tag \}\\[0.8em]

\textbf{Requirements:}
\begin{enumerate}[leftmargin=1.6em,itemsep=0.2em,topsep=0.2em]
  \item Do not add, delete, or paraphrase any words in the original prompt except inserting the bbox tags.
  \item Place the bbox tag immediately after the exact subject (e.g., ``a boy'' $\rightarrow$ ``a boy\textless bbox\textgreater[x1,y1,x2,y2]\textless/bbox\textgreater{} from image N'').
  \item Return only the modified prompt (no extra explanations or sentences).
  \item Insert the tag only after the \textbf{first} occurrence of each subject (if duplicated).
\end{enumerate}

\vspace{0.6em}
\textbf{Example:}\\[0.2em]
\textit{Original:}\\
``The cat plays with the yarn ball on a sunny windowsill.''\\[0.2em]
\textit{Tags:}\\
``cat'' $\rightarrow$ \texttt{<bbox>[0.1,0.2,0.3,0.4]</bbox>} from image1;\\
``yarn ball'' $\rightarrow$ \texttt{<bbox>[0.5,0.6,0.7,0.8]</bbox>} from image2\\[0.2em]
\textit{Modified:}\\
``The cat\textless bbox\textgreater[0.1,0.2,0.3,0.4]\textless/bbox\textgreater{} from image1 plays with the yarn ball\textless bbox\textgreater[0.5,0.6,0.7,0.8]\textless/bbox\textgreater{} from image2 on a sunny windowsill.''

\end{tcolorbox}
\caption{Prompt template for inserting \texttt{<bbox>} tags into $p^{\text{raw}}_{\text{dense}}$ (Reference-conditioned Branch)}
\label{fig:bbox_insertion_prompt}
\end{figure*}

\end{document}